\documentclass{article}

\usepackage{microtype}
\usepackage{graphicx}
\usepackage{subfigure}
\usepackage{booktabs}
\usepackage{multirow}
\usepackage{hyperref}
\usepackage{MnSymbol}

\usepackage{color}
\usepackage[accepted]{icml2024}

\usepackage{amsmath}
\usepackage{mathtools}
\usepackage{amsthm}
\usepackage{mathrsfs}

\usepackage[capitalize,noabbrev]{cleveref}
\definecolor{antiquebrass}{rgb}{0.8, 0.58, 0.46}
\definecolor{amber}{rgb}{1.0, 0.49, 0.0}
\definecolor{apricot}{rgb}{0.98, 0.81, 0.69}
\definecolor{bronze}{rgb}{0.8, 0.5, 0.2}
\hypersetup{
    colorlinks=true,
    linkcolor=red,
    urlcolor=bronze,
    pdfpagemode=FullScreen,
    }

\theoremstyle{plain}

\theoremstyle{definition}

\theoremstyle{remark}

\usepackage[textsize=tiny]{todonotes}
\usepackage{amsfonts}
\usepackage{listings}
\usepackage{multirow} 
\usepackage{enumitem}
\usepackage{color,soul}
\usepackage{color, colortbl}
\definecolor{Gray}{gray}{0.93}
\definecolor{Orange}{rgb}{1,0.5,0}
\definecolor{DGray}{gray}{0.9}
\definecolor{LightCyan}{rgb}{0.88,1,1}
\icmltitlerunning{Parameter-Efficient Fine-Tuning with Discrete Fourier Transform}

\begin{document}
\newcommand{\highlight}[1]{\colorbox{blue!10}{#1}}
\newcommand{\stardiy}{\vcenter{\hbox{\includegraphics[scale=0.06]{figure/star.pdf}}}}
\newcommand{\tridiy}{\vcenter{\hbox{\includegraphics[scale=0.06]{figure/tri.pdf}}}}
\twocolumn[
\icmltitle{
Parameter-Efficient Fine-Tuning with Discrete Fourier Transform
}



\icmlsetsymbol{equal}{*}

\begin{icmlauthorlist}
\icmlauthor{Ziqi Gao}{hkustgz,hkust,equal}
\icmlauthor{Qichao Wang}{sysu,equal}
\icmlauthor{Aochuan Chen}{hkustgz,equal}
\icmlauthor{Zijing Liu}{idea}
\icmlauthor{Bingzhe Wu}{tencent}
\icmlauthor{Liang Chen}{sysu}
\icmlauthor{Jia Li}{hkustgz,hkust}

\end{icmlauthorlist}

\icmlaffiliation{hkustgz}{Hong Kong University of Science and Technology
(Guangzhou)}
\icmlaffiliation{hkust}{Hong Kong University of Science and Technology}
\icmlaffiliation{sysu}{Sun Yat-sen University}
\icmlaffiliation{idea}{International Digital Economy Academy}
\icmlaffiliation{tencent}{AI Lab, Tencent}
\icmlcorrespondingauthor{Jia Li}{jialee@ust.hk}

\icmlkeywords{Machine Learning, ICML}

\vskip 0.3in
]



\makeatletter\def\Hy@Warning#1{}\makeatother

\printAffiliationsAndNotice{\icmlEqualContribution} 

\begin{abstract}
Low-rank adaptation~(LoRA) has recently gained much interest in fine-tuning foundation models. It effectively reduces the number of trainable parameters by incorporating low-rank matrices $A$ and $B$ to represent the weight change, i.e., $\Delta W=BA$. Despite LoRA's progress, it faces storage challenges when handling extensive customization adaptations or larger base models. In this work, we aim to further compress trainable parameters by enjoying the powerful expressiveness of the Fourier transform. Specifically, we introduce FourierFT, which treats $\Delta W$ as a matrix in the spatial domain and learns only a small fraction of its spectral coefficients. With the trained spectral coefficients, we implement the inverse discrete Fourier transform to recover $\Delta W$. Empirically, our FourierFT method shows comparable or better performance with fewer parameters than LoRA on various tasks, including natural language understanding, natural language generation, instruction tuning, and image classification. For example, when performing instruction tuning on the LLaMA2-7B model, FourierFT surpasses LoRA with only 0.064M trainable parameters, compared to LoRA's 33.5M. Our code is released at \url{https://github.com/Chaos96/fourierft}.
\end{abstract}
\section{Introduction}
Large foundation models~(LFMs) have demonstrated exceptional performance on tasks of multiple domains, including natural language processing (NLP)~\cite{roberta,deberta,gpt2,gpt3,licommunity} and computer vision (CV)~\cite{cv_2,cv_1,cv_3,sd}. Owing to their impressive capabilities, fine-tuning LFMs for a wide range of downstream tasks has become prevalent~\cite{gpt35,alpaca,ft_3}. Under the full fine-tuning paradigm, the new model adapted to each customized task typically contains as many parameters as the original model~\cite{ft_3,ft_4,graphwiz,promptmsp}. As models grow larger and customization needs expand, the demand for storing fine-tuned checkpoints rises, resulting in both costly storage and memory consumption.

\begin{figure}[t]
\centering
\includegraphics[width=0.47\textwidth]{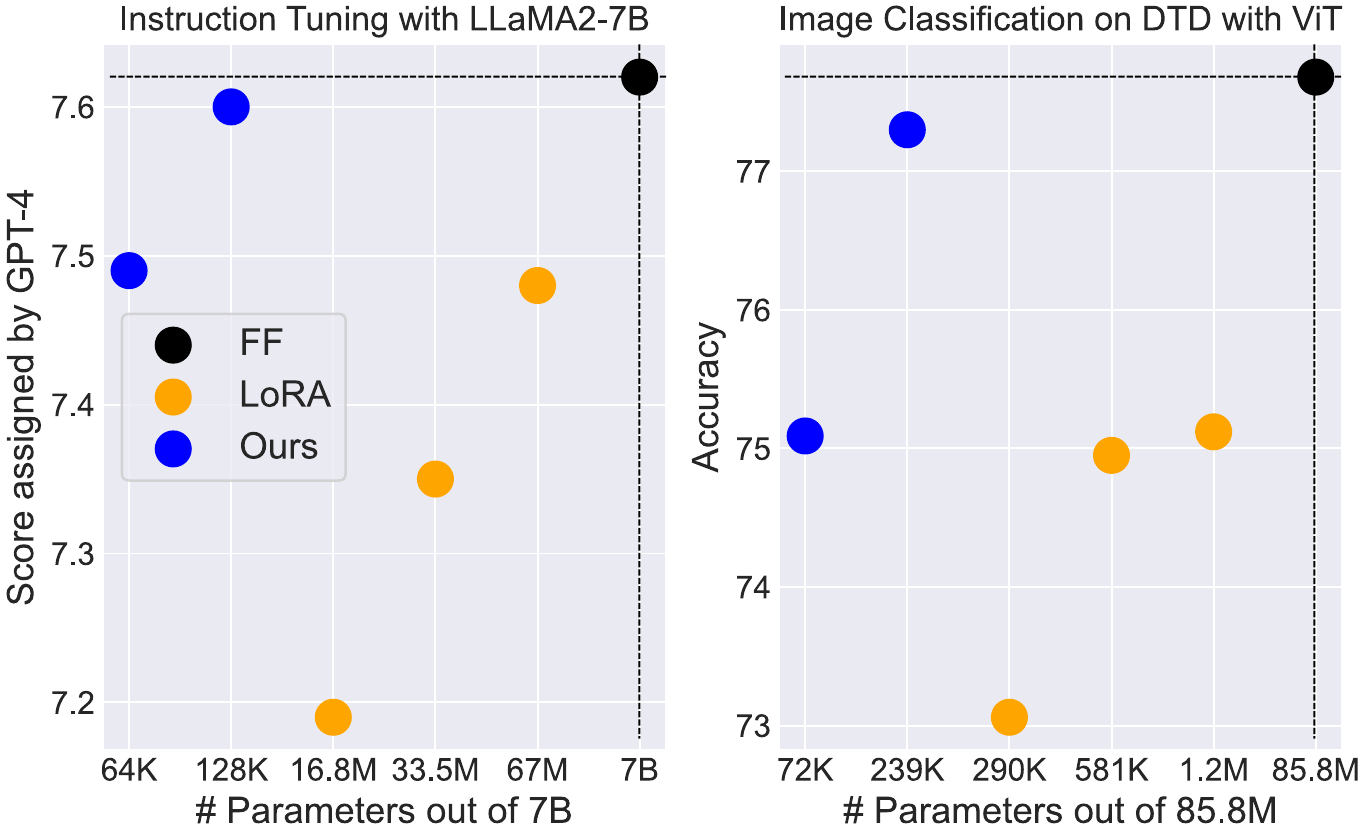} 
\vspace{-0.2em}
 \caption{Summary of the performance (\textbf{y-axis}) of fine-tuning methods with different numbers (\textbf{x-axis}) of trainable parameters on NLP (left) and CV (right) tasks. The left side shows the instruction tuning task, where the LLaMA2-7B model is fine-tuned with Alpaca and evaluated by GPT-4. The right side shows the image classification task, where the Vision Transformer (ViT) is fine-tuned and tested on the DTD dataset.
\textit{Black circles}~($\color{black}{\bullet}$) represent the Full Fine-tuning (FF) method. \textit{Orange circles}~($\color{orange}{\bullet}$) represent LoRA method with $r=\{32,64,128\}$ (left) and $r=\{8,16,32\}$ (right).  \textit{Blue circles}~($\color{blue}{\bullet}$) represent our proposed method with $n=\{1000, 2000\}$ (left) and $n=\{3000, 10000\}$ (right).}
\label{fig:motivation}
\end{figure}

As a popular way to address this issue, LoRA \cite{lora} represents the weight change with two low-rank matrices $A$ and $B$, i.e., $W_0+\Delta W = W_0+BA$.
Despite LoRA's superb performance, its large size of trainable parameters still brings high IT infrastructure consumption, which affects both ends of public communities and individual users. For the former, an intuitive example is that a LoRA adapter~(fine-tuned weights) for a specific style of the stable diffusion model~\cite{sd} requires about 40MB of memory. This necessitates the LFM communities (e.g., Civitai~\cite{civitai}) to bear high storage and bandwidth costs to cater to a large user base. For the latter, fewer parameters mean direct RAM savings when loading fine-tuned weights in mobile APPs, enabling sufficient customization for individual users~\cite{pets1}. To this end, we naturally ask the question: \textit{How can we aggressively compress trainable parameters even further for fine-tuning LFMs?}

Previous works have demonstrated the powerful expressiveness of Fourier basis in data compression, where extremely sparse spectral information can be used to recover high-fidelity data~(e.g., 1D signal vectors \cite{fft_signal_1,fft_signal_2,fft_signal_3} and 2D image matrices \cite{fft_cv_1,fft_cv_2,fft_cv_3}). More importantly, when dealing with more general (non-image) matrices that lack strong spatial semantics and are not frequency-sparse, Fourier transform can still handle recovery effectively \cite{evi_2,evi_3}. Motivated by this, we investigate the potential for updating the weight change $\Delta W$ with its sparse spectral coefficients for fine-tuning LFMs. 

In this paper, we aim to aggressively reduce the number of trainable parameters for fine-tuning LFMs. To this end, we propose \textit{FourierFT} (\underline{Fourier} Transform for \underline{F}ine-\underline{T}uning), which treats the weight change $\Delta W$ as a matrix in the spatial domain, and learns its sparse spectral coefficients. Specifically, we first randomly select $n$ spectral entries that are shared across all layers. For each layer, FourierFT learns $n$ spectral coefficients located at these $n$ selected entries and then directly applies inverse discrete Fourier transform to compute the updated $\Delta W$. Therefore, fine-tuning a pre-trained model with $L_{t}$ layers only requires storing $2n$ entry parameters and $nL_{t}$ coefficient parameters for FourierFT. 

Empirically, we compare our method with state-of-the-art LoRA variants and other parameter-efficient fine-tuning methods on various tasks including (1) natural language understanding~(on the GLUE benchmark), (2) natural language generation~(on the E2E benchmark), (3) instruction tuning (with LLaMA-family models), and (4) image classification~(with vision transformers). FourierFT can always achieve comparable or even better performance than LoRA, with about 6.0\%, 9.4\%, 0.2\% and 9.2\% of LoRA's trainable parameters for these 4 tasks, respectively. For example in Figure \ref{fig:motivation}, on the instruction tuning task, our FourierFT method outperforms LoRA with only 64K trainable parameters. Moreover, it achieves a comparable score to Full Fine-tuning with only 128K parameters.
\section{Related Works}
\paragraph{Parameter-Efficient Fine-Tuning.} 
With the rapid expansion of large foundation models (LFM), it has become challenging and important to efficiently adapt them for specific tasks. To this end, numerous methods for parameter-efficient fine-tuning (PEFT) are proposed, demonstrating impressive capabilities in both efficiency and accuracy. Existing PEFT methods are broadly partitioned into two categories: non-weight-based and weight-based methods.

\textbf{Non-weight-based methods} do not optimize pre-trained LFMs at the weight level. Instead, they achieve fine-tunings by introducing additional modules or optimizing prompts and prefixes. Adapter tuning~\cite{adapter1,adapter2,ada_p,ada_h,ada_d,ada_l} aims to introduce light-weighted neural modules, called adapters, between pre-trained layers of the base model. These methods keep the pre-trained weights frozen and efficiently fine-tune the adapters for customized tasks. Prompt tuning~\cite{gpt3,prompt1,prompt2,prompt3} and prefix tuning~\cite{prefix1} insert additional prompts or prefix tokens to the layers of the base model.
\textbf{Weight-based methods}, represented by LoRA~\cite{lora}, introduce and then update weight changes that can be merged with the original weights to avoid inference latency. LoRA’s innovation lies in the multiplication of low-rank matrices to approximate weight changes. Building upon this, AdaLoRA~\cite{adalora} extends the LoRA method by distributing the parameter budget across weight matrices with importance scores. Additionally, Q-LoRA~\cite{qlora} proposes to back-propagate gradients upon LoRA through a quantized pre-trained model with 4-bit NormalFloat.

Here, we focus on \textbf{weight-based methods} and achieve huge parameter reduction with the powerful expressiveness of Fourier basis, rather than following the low-rank structure.

\paragraph{Sparse Fourier Transform in Deep Learning.} Sparse Fourier transform~(SFT) has flourished in various fields of deep learning~(DL). The SFT technique mainly involves using sparse spectral coefficients of significant \cite{data3,data1,data2,bwgnn} or even random~\cite{random1,random2,random3} spectral entries, for representation learning. One important application of this technique is matrix recovery. \citet{recover1} designs a gradient-based compressed sensing method to recover images with their sparse Fourier information. \citet{recover2} proposes an efficient phase retrieval method that improves data recovery using sparse Fourier coefficients. Importantly, previous works~\cite{evi_2,evi_3,megae} show that even when the original data is not frequency-sparse, SFT can effectively recover the data with extremely few parameters. Although previous works lack studies on the recovery for the weight matrices of DL models with SFT, the aforementioned methods provide potential support for this work.


\begin{figure*}[t]
\centering
\includegraphics[width=0.8\textwidth]{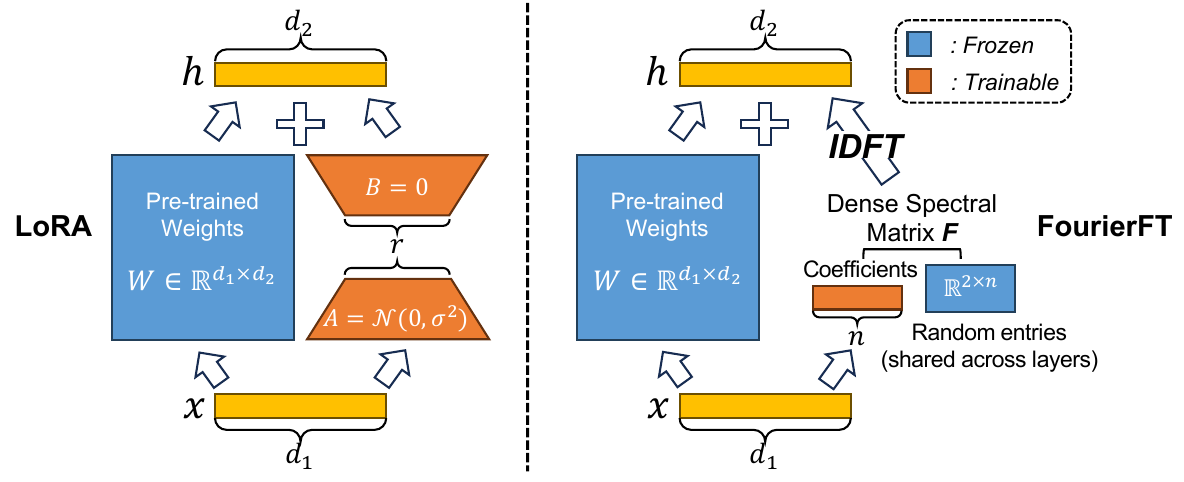} 
\caption{\textbf{Overview of LoRA~(left) and our FourierFT~(right) method.} In LoRA, only low-rank~($r$) matrices $A$ and $B$ are trained. The weight change is represented by their multiplication, i.e., $\Delta W=BA$. For each pre-trained weight $W$, the theoretical number of trainable parameters in LoRA is $r\times(d_{1}+d_{2})$. In FourierFT, we first randomly generate the spectral entry matrix $\mathbb{R}^{2\times n}$, which is shared across all layers to reduce parameter storage requirements. The complete spectral matrix is formed by a trainable coefficient vector $\mathbb{R}^n$ located at selected entries and $0$s at the remaining entries. We obtain the weight change $\Delta W$ by directly performing inverse discrete Fourier transform (IDFT) on the updated spectral matrix. For all $L$ adapted layers, FourierFT needs to store $n\times(2+L)$ parameters.}
\label{fig:model}
\end{figure*}
\section{Method}
We present FourierFT~(depicted in Figure~\ref{fig:model}), a parameter-efficient fine-tuning method based on discrete Fourier transform. FourierFT follows the principle of only learning the change in the pre-trained weight, as proposed by LoRA~\cite{lora}. However, unlike LoRA, FourierFT does not adopt the low-rank structure but learns a set of spectral coefficients of Fourier basis. Specifically, we randomly initialize the spectral entry matrix, which is frozen and shared across all layers. We make the spectral coefficients located at selected entries trainable, which jointly form the spectral matrix. Lastly, we apply the inverse discrete Fourier transform to the spectral matrix, yielding its spatial-domain counterpart as the updated weight change.
\subsection{Forward Pass}
We follow the paradigm of only learning weight changes, as adopted by LoRA-based methods~\cite{lora,qlora,adalora}. This can avoid inference latency by merging the pre-trained weight and its change. Formally, we define each pre-trained weight matrix as $W_0\in\mathbb{R}^{d_1\times d_2}$, and the weight change for fine-tuning as $\Delta W\in\mathbb{R}^{d_1\times d_2}$. LoRA aims to parameterize $\Delta W$ in the form of low-rank decomposition in the forward pass:
\begin{equation}
    h=W_0x+\Delta Wx=W_0x+BAx,
\end{equation}
where $B\in\mathbb{R}^{d_1\times r}$ and $A\in\mathbb{R}^{r\times d_2}$ with the rank $r\ll \text{min}(d_1,d_2)$ are trainable matrices.

The advantage of FourierFT is that the orthogonal and expressive Fourier basis enables recovery of informative weight changes. This promisingly suggests achieving comparable performance to LoRA with significantly fewer parameters. We first randomly initialize the entry matrix $E\in\mathbb{R}^{2\times n}$ containing discrete 2D spectral entries. Then we randomly initialize the coefficients $c\in\mathbb{R}^n$ with a normal Gaussian distribution. The proposed forward pass is:
\begin{equation}\label{eq2}
    F=\textsc{ToDense}(E,c)
\end{equation}
\begin{equation}\label{eq3}
    S_{p,q} = \sum_{j=0}^{d_1-1}\sum_{k=0}^{d_2-1}F_{j,k}e^{i2\pi\left( \frac{p}{d_1}j + \frac{q}{d_2}k  \right)}
\end{equation}
\begin{equation}\label{eq4}
\begin{aligned}
        h & =W_0x+\Delta Wx \\& =W_0x+\alpha \Re(S) x.
\end{aligned}
\end{equation}
Specifically, \textsc{ToDense} in Eq.~\ref{eq2} represents to construct the \textbf{spectral matrix} $F\in\mathbb{R}^{d_1\times d_2}$, i.e., $F_{j,k}=c_l$ (resp. 0), if $j=E_{0,l}$ \& $k=E_{1,l}$ (resp. else). Eq.~\ref{eq3} computes the \textbf{spatio matrix} $S$ via the inverse discrete Fourier transform, where $i$ represents the imaginary unit. Finally, in Eq.~\ref{eq4}, we take the real part of the complex matrix $S$ (denoted as $\Re(S)$) and scale it by $\alpha$. Kindly note that all layers involve training various $c$ vectors, while sharing the matrix $E$ and value $\alpha$.

\begin{figure}[ht]
\centering
\includegraphics[width=0.48\textwidth]{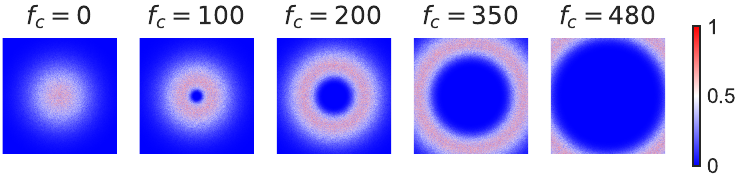} 
\vspace{-1.5em}
 \caption{Visualization of entry sampling probability at different favored central frequencies $f_c$.}
\label{fig:bias}
\end{figure}

The pseudocode for FourierFT is shown as Algorithm~\ref{algo:main}, adhering to the PyTorch style. 
\begin{algorithm}[t]
   \caption{PyTorch-style pseudocode for \textbf{FourierFT}.}
   \label{algo:main}
   
    \definecolor{codeblue}{rgb}{0.25,0.5,0.5}
    \lstset{
      basicstyle=\fontsize{7.2pt}{7.2pt}\ttfamily\bfseries,
      commentstyle=\fontsize{7.2pt}{7.2pt}\color{codeblue},
      keywordstyle=\fontsize{7.2pt}{7.2pt},
    }
\begin{lstlisting}[language=python]
class FourierFT(nn.Module):
    def __init__(
    self,
    n: int = 100, # number of trainable parameters 
    alpha: float = 300.0, # scaling 
    d1: int = 4096, # input dimension
    d2: int = 4096, # output dimension
    base_layer: nn.Module # pre-trained layer
    )
        # definitions
        self.d1 = d1
        self.d2 = d2
        self.n = n
        self.alpha = alpha
        self.base_layer = base_layer
        # entry initialization (no frequency bias)
        self.E = torch.randperm(d1 * d2)[:n]
        self.E = torch.stack([self.E // self.d1,
        self.E % self.d2], dim=0)
        # spectral coefficient initialization
        self.c = nn.Parameter(torch.randn(n), \\
            requires_grad=True)

    def forward(self, x: torch.Tensor):
        # get dense spectral matrix (Eq.2)
        F = torch.zeros(self.d1, self.d2)
        F[self.E[0, :], self.E[1, :]] = self.c
        # compute Delta_W (Eq.3)
        Delta_W = torch.fft.ifft2(F).real * self.alpha
        # merge (Eq.4)
        h = self.base_layer(x)
        h += torch.einsum('ijk,kl->ijl', x, Delta_W)
        return h

\end{lstlisting}
\end{algorithm}
\paragraph{Initialization for the Entry Matrix $E$.}Previous works lack studies on the importance of the spectral entries in the weight change. Thus, we fill this gap by introducing adjustable frequency bias, causing the entries to be more likely sampled in this area. In addition to randomly sampling entries in the full $d_1\times d_2$-sized spectral matrix (i.e., no bias), we also implement entry sampling with a bias towards a favored central frequency, e.g., low, middle, or high frequencies.
Formally, we apply the Gaussian band-pass filter~\cite{filter} to model the sampling probability for the entry $(u,v), 0\leq u\leq d_1-1, 0\leq v\leq d_2-1$:
\begin{equation}\label{eq5}
    p(u,v) = \exp\left({-\left(\frac{\mathcal{D}^2-f_c^2}{\mathcal{D}\mathcal{W}}\right)^2}\right),
\end{equation}
where $\mathcal{D}$ represents the distance from the point $(u, v)$ to the origin (center of the matrix), $f_c$ is the favored central frequency, and $\mathcal{W}$ represents the bandwidth. In Figure~\ref{fig:bias}, we visualize the sampling probability map of a $768\times768$-sized spectral matrix with different $f_c$ and $\mathcal{W}=200$. 

Kindly note that unless specially stated, FourierFT is set by default to the entry initialization with no frequency bias.


\subsection{Parameter Summary}\label{sec32}
\begin{table}[t]
\centering
\caption{Theoretical number of trainable parameters and storage requirements for fine-tuning. For both LoRA and FourierFT methods, only the query and value layers are tuned within the transformer architectures. The configurations that are exactly chosen in the `Experiments' Section are \highlight{highlighted}.}
\label{tab:para}
\resizebox{0.48\textwidth}{!}{%
\begin{tabular}{@{}c|ccc|ccc@{}}
\toprule[1.1pt]
\multirow{2}{*}{\textbf{Base Models}} & \multicolumn{3}{c|}{\textbf{LoRA}} & \multicolumn{3}{c}{\textbf{FourierFT}} \\ \cmidrule(l){2-7} 
 & $r$ & \begin{tabular}[c]{@{}c@{}}\# Trainable\\ Parameters\end{tabular} & \begin{tabular}[c]{@{}c@{}}Required\\ Bytes\end{tabular} & $n$ & \begin{tabular}[c]{@{}c@{}}\# Trainable\\ Parameters\end{tabular} & \begin{tabular}[c]{@{}c@{}}Required\\ Bytes\end{tabular} \\ \midrule[0.8pt]
\multirow{2}{*}{\begin{tabular}[c]{@{}c@{}}RoBERTa\\ Base\end{tabular}} & 4 & 147K & 574KB & 200 & 4.8K & 18.8KB \\
 & \highlight{8} & \highlight{295K} & \highlight{1.13MB} & \highlight{200} & \highlight{24K} & \highlight{94KB} \\ \midrule
\multirow{2}{*}{\begin{tabular}[c]{@{}c@{}}RoBERTa\\ Large\end{tabular}} & 4 & 393K & 1.5MB & 200 & 9.6K & 36.5KB \\
 & \highlight{8} & \highlight{786K} & \highlight{3MB} & \highlight{1000} & \highlight{48K} & \highlight{183KB} \\ \midrule
\multirow{2}{*}{\begin{tabular}[c]{@{}c@{}}GPT-2\\ Medium\end{tabular}} & \highlight{4} & \highlight{350K} & \highlight{1.34MB} & 500 & 24K & 94KB \\
 & 8 & 786K & 3MB & \highlight{1000} & \highlight{48K} & \highlight{188KB} \\ \midrule
\multirow{2}{*}{\begin{tabular}[c]{@{}c@{}}GPT-2\\ Large\end{tabular}} & \highlight{4} & \highlight{737K} & \highlight{2.81MB} & 500 & 36K & 141KB \\
 & 8 & 1.47M & 5.74MB & \highlight{1000} & \highlight{72K} & \highlight{282KB} \\ \midrule
\multirow{2}{*}{\begin{tabular}[c]{@{}c@{}}LLaMA-2\\ 7B\end{tabular}} & 16 & 8.39M & 32.8MB & \highlight{1000} & \highlight{64K} & \highlight{250KB} \\
 & \highlight{64} & \highlight{33.5M} & \highlight{131.1MB} & 2000 & 128K & 500KB \\ \midrule
\multirow{2}{*}{\begin{tabular}[c]{@{}c@{}}LLaMA-2\\ 13B\end{tabular}} & 16 & 13.1M & 51.2MB & \highlight{1000} & \highlight{80K} & \highlight{312KB} \\
 & \highlight{64} & \highlight{52.4M} & \highlight{204.8MB} & 2000 & 160K & 625KB \\ \midrule
\multirow{2}{*}{\begin{tabular}[c]{@{}c@{}}ViT\\ Base\end{tabular}} & 8 & 295K & 1.13MB & \highlight{3000} & \highlight{72K} & \highlight{281KB} \\
 & \highlight{16} & \highlight{590K} & \highlight{2.25MB} & 10000 & 239K & 934KB \\
 \midrule
\multirow{2}{*}{\begin{tabular}[c]{@{}c@{}}ViT\\ Large\end{tabular}} & 8 & 786K & 2.93MB & \highlight{3000} & \highlight{144K} & \highlight{563KB} \\
 & \highlight{16} & \highlight{1.57M} & \highlight{6MB} & 10000 & 480K & 1.83MB \\\bottomrule[1pt]
\end{tabular}%
}
\end{table}
We summarize the number of trainable parameters for LoRA and FourierFT in Table~\ref{tab:para}. LoRA relies on a pair of trainable matrices $A$ and $B$ for each layer. Let the number of layers for fine-tuning be $L_t$. The total number of parameters in LoRA is determined by the rank $r$ and the dimension of weights $d=d_1=d_2$: $\left| \Theta \right|_{LoRA}=2\times d\times L_t\times r$. For Fourier, the total number takes the form: $\left| \Theta \right|_{FourierFT}=n\times L_t$. As an intuitive example, the RoBERTa Base model contains 12 transformer blocks with $d=768$, resulting in $L_t=24$ layers when we only fine-tune the query and value ones. Therefore, we have $\left| \Theta \right|_{LoRA}=294,912$ for $r=8$, and $\left| \Theta \right|_{FourierFT}=24,000$ for $n=1000$. In Table~\ref{tab:para}, we \highlight{highlight} the configurations where LoRA and our method achieve matched performance in subsequent experiments. We note that the advantage of parameter efficiency in FourierFT becomes more pronounced as the model's scale~(depth and width) increases (e.g., RoBERTa Base $\rightarrow$ RoBERTa Large). This could be because $\left| \Theta \right|_{LoRA}$ has an explicit linear relationship with width $d$, unlike $\left| \Theta \right|_{FourierFT}$. 

\section{Experiments}
\begin{table*}[t]
\centering
\caption{Performance of various fine-tuning methods with RoBERTa Base~($\rm{RoB_{base}}$) and RoBERTa Large~($\rm{RoB_{large}}$) models on 6 datasets of the GLUE benchmark. We report the Matthew's correlation coefficient~(MCC) for CoLA, Pearson correlation coefficient~(PCC) for STS-B and accuracy~(Acc.) for all the remaining tasks. We report the median result of 5 runs, each using different random seeds. The best results for each dataset are shown in \textbf{bold}. Higher is better for all metrics in 6 datasets.}
\label{tab:glue}
\addtolength{\tabcolsep}{-1pt}
\resizebox{0.9\textwidth}{!}{%
\begin{tabular}{@{}l|r|cccccccc@{}}
\toprule
\textbf{Model \& Method} & \multicolumn{1}{c|}{\begin{tabular}[c]{@{}c@{}}\# Trainable\\ Parameters\end{tabular}} & \begin{tabular}[c]{@{}c@{}}\textbf{SST-2}\\ (Acc.)\end{tabular} & \begin{tabular}[c]{@{}c@{}}\textbf{MRPC}\\ (Acc.)\end{tabular} & \begin{tabular}[c]{@{}c@{}}\textbf{CoLA}\\ (MCC)\end{tabular} & \begin{tabular}[c]{@{}c@{}}\textbf{QNLI}\\ (Acc.)\end{tabular} &  \begin{tabular}[c]{@{}c@{}}\textbf{RTE}\\ (Acc.)\end{tabular} & \begin{tabular}[c]{@{}c@{}}\textbf{STS-B}\\ (PCC)\end{tabular} & \multicolumn{1}{c}{\textbf{Avg.}} \\ \midrule
$\rm{RoB_{base}}$(FF) & 125M &  94.8 & 90.2 & \textbf{63.6} & 92.8  & 78.7 & 91.2 & 85.2 \\
$\rm{RoB_{base}}$(BitFit) & 0.1M  & 93.7 & \textbf{92.7} & 62 & 91.8 & \textbf{81.5}& 90.8 & \textbf{85.4} \\
$\rm{RoB_{base}}$($\text{Adpt}^{\text{D}}$) & 0.3M &  94.2\textsubscript{$\pm$0.1} & 88.5\textsubscript{$\pm$1.1} & 60.8\textsubscript{$\pm$0.4} & 93.1\textsubscript{$\pm$0.1} & 71.5\textsubscript{$\pm$2.7} & 89.7\textsubscript{$\pm$0.3} & 83.0 \\
$\rm{RoB_{base}}$($\text{Adpt}^{\text{D}}$) & 0.9M & 94.7\textsubscript{$\pm$0.3} & 88.4\textsubscript{$\pm$0.1} & 62.6\textsubscript{$\pm$0.9} & 93.0\textsubscript{$\pm$0.2} & 75.9\textsubscript{$\pm$2.2} & 90.3\textsubscript{$\pm$0.1} & 84.2 \\
$\rm{RoB_{base}}$(LoRA) & 0.3M & \textbf{95.1}\textsubscript{$\pm$0.2} & 89.7\textsubscript{$\pm$0.7} & 63.4\textsubscript{$\pm$1.2} & \textbf{93.3}\textsubscript{$\pm$0.3} & 78.4\textsubscript{$\pm$0.8} & \textbf{91.5}\textsubscript{$\pm$0.2} & 85.2 \\
$\rm{RoB_{base}}$(AdaLoRA) & 0.3M & 94.5\textsubscript{$\pm$0.2} & 88.7\textsubscript{$\pm$0.5} & 62.0\textsubscript{$\pm$0.6} & 93.1\textsubscript{$\pm$0.2} & 81.0\textsubscript{$\pm$0.6} & 90.5\textsubscript{$\pm$0.2} & 85.0 \\
$\rm{RoB_{base}}$(DyLoRA) & 0.3M & 94.3\textsubscript{$\pm$0.5} & 89.5\textsubscript{$\pm$0.5} & 61.1\textsubscript{$\pm$0.3} & 92.2\textsubscript{$\pm$0.5} & 78.7\textsubscript{$\pm$0.7} & 91.1\textsubscript{$\pm$0.6} & 84.5 \\
\textbf{$\rm{RoB_{base}}$(FourierFT)} & 0.024M & 94.2\textsubscript{$\pm$0.3} & 90.0\textsubscript{$\pm$0.8} & \textbf{63.8}\textsubscript{$\pm$1.6} & 92.2\textsubscript{$\pm$0.1} & 79.1\textsubscript{$\pm$0.5} & 90.8\textsubscript{$\pm$0.2} & 85.0 \\ \midrule[0.6pt]
$\rm{RoB_{large}}$(FF) & 356M &  96.4 & \textbf{90.9} & 68 & 94.7 & 86.6 & \textbf{92.4 }& \textbf{88.2} \\
$\rm{RoB_{large}}$($\text{Adpt}^{\text{P}}$) & 3M & 96.1\textsubscript{$\pm$0.3} & 90.2\textsubscript{$\pm$0.7} & \textbf{68.3}\textsubscript{$\pm$1.0} & \textbf{94.8}\textsubscript{$\pm$0.2} & 83.8\textsubscript{$\pm$2.9} & 92.1\textsubscript{$\pm$0.7} & 87.6 \\
$\rm{RoB_{large}}$($\text{Adpt}^{\text{P}}$) & 0.8M & \textbf{96.6}\textsubscript{$\pm$0.2} & 89.7\textsubscript{$\pm$1.2} & 67.8\textsubscript{$\pm$2.5} & \textbf{94.8}\textsubscript{$\pm$0.3} & 80.1\textsubscript{$\pm$2.9} & 91.9\textsubscript{$\pm$0.4} & 86.8 \\
$\rm{RoB_{large}}$($\text{Adpt}^{\text{H}}$) & 6M & 96.2\textsubscript{$\pm$0.3} & 88.7\textsubscript{$\pm$2.9} & 66.5\textsubscript{$\pm$4.4} & 94.7\textsubscript{$\pm$0.2} &  83.4\textsubscript{$\pm$1.1} & 91.0\textsubscript{$\pm$1.7} & 86.8 \\
$\rm{RoB_{large}}$($\text{Adpt}^{\text{H}}$) & 0.8M & 96.3\textsubscript{$\pm$0.5} & 87.7\textsubscript{$\pm$1.7} & 66.3\textsubscript{$\pm$2.0} & 94.7\textsubscript{$\pm$0.2} & 72.9\textsubscript{$\pm$2.9} & 91.5\textsubscript{$\pm$0.5} & 84.9 \\
$\rm{RoB_{large}}$(LoRA) & 0.8M & 96.2\textsubscript{$\pm$0.5} & 90.2\textsubscript{$\pm$1.0} & 68.2\textsubscript{$\pm$1.9} & \textbf{94.8}\textsubscript{$\pm$0.3} & 85.2\textsubscript{$\pm$1.1} & 92.3\textsubscript{$\pm$0.5} & 87.8 \\
$\rm{RoB_{large}}$(\textbf{FourierFT}) & 0.048M & 96.0\textsubscript{$\pm$0.2} & \textbf{90.9}\textsubscript{$\pm$0.3} & 67.1\textsubscript{$\pm$1.4} & 94.4\textsubscript{$\pm$0.4} & \textbf{87.4}\textsubscript{$\pm$1.6} & 91.9\textsubscript{$\pm$0.4} & 88.0 \\ \bottomrule
\end{tabular}%
}
\end{table*}
In this section, we evaluate FourierFT in the domains of natural language processing~(NLP) and computer vision~(CV). For NLP, we implement FourierFT for fine-tuning \textbf{(1)} RoBERTa (Base \& Large) on natural language understanding~(GLUE,~\cite{glue}), \textbf{(2)} GPT-2~(Medium \& Large) on natural language generation~(E2E,~\cite{e2e}) and \textbf{(3)} LLaMA-family models~(7B \& 13B) on instruction tuning. For CV, we apply FourierFT to fine-tune the \textbf{(4)} vision transformers (Base \& Large) on image classification. Finally, we conduct ablation studies to analyze the effect of frequency bias, the parameter scalability, and the expressiveness of the Fourier basis.
\paragraph{Baselines.}We compare our FourierFT method with popular parameter-efficient fine-tuning (PEFT) methods. To ensure a comprehensive and fair comparison, we prioritize replicating the setups used in previous works and reusing their reported results. Involved baselines are:\\
$ \bullet $ \textbf{Full Fine-tuning (FF)} - During fine-tuning, the base model is initialized with pre-trained weights and biases, and all parameters will undergo gradient updates.\\
$ \bullet $ \textbf{Bitfit}~\cite{bitfit} - Only the bias vectors are fine-tuned while all other parameters are frozen.\\
$ \bullet $ \textbf{Adapter tuning} - This research line was first investigated by~\citet{ada_h}, which proposes the $\textbf{Adapter}^{\textbf{H}}$ method. $\textbf{Adapter}^{\textbf{H}}$ inserts two-layer adapters between the self-attention and the FNN modules, followed by a subsequent residual connection. We compare it with three additional variants of it. $\textbf{Adapter}^{\textbf{L}}$~\cite{ada_l} is more parameter-efficient, with adapter layers applied only after the MLP modules and subsequent to a LayerNorm. $\textbf{Adapter}^{\textbf{P}}$~\cite{ada_p} implements the adapter layers after the feed-forward layer. This design was chosen through a grid search including all settings related to the adapter's position, number, \textit{ect}. $\textbf{Adapter}^{\textbf{D}}$~\cite{ada_d} further enhances the parameter efficiency by dropping adapter layers that are not activated.\\
$ \bullet $ \textbf{LoRA}~\cite{lora} - LoRA is the state-of-the-art method for PEFT. It parameterizes incremental weight updates using trainable low-rank matrices.\\
$ \bullet $ \textbf{DyLoRA}~\cite{dylora} - This method trains dynamic search-free LoRA models for the best rank choice.
$ \bullet $ \textbf{AdaLoRA}~\cite{adalora} - This method proposes the SVD-based fine-tuning and prunes redundant singular values with the importance-aware rank allocation. 
\begin{table*}[t]
 \begin{minipage}[t]{0.58\textwidth}
  \caption{Results from GPT-2 Medium and Large models on the E2E benchmark. We present the result from the final epoch. For all metrics, higher values indicate better performance. * indicates that the results are taken from prior works. Best results are shown in \textbf{bold}. }
\label{tab:e2e}
\addtolength{\tabcolsep}{-4.3pt}
  \resizebox{\textwidth}{!}{%
\begin{tabular}{@{}c|l|r|ccccc@{}}
\toprule
Model & Method & \multicolumn{1}{c|}{\begin{tabular}[c]{@{}c@{}}\# Trainable\\ Parameters\end{tabular}} & BLEU & NIST & METEOR & ROUGE-L & CIDEr \\ \midrule
\multirow{6}{*}{\rotatebox{90}{\begin{tabular}[c]{@{}c@{}} GPT-2\\ Medium\end{tabular}}} & FT*        & 354.92M & 68.2 & 8.62 & 46.2 & 71.0 & 2.47 \\
& \multicolumn{1}{l|}{$\text{Adpt}^{\text{L}}$*}         & \multicolumn{1}{r|}{0.37M}   & 66.3 & 8.41 & 45.0 & 69.8 & 2.40 \\
& \multicolumn{1}{l|}{$\text{Adpt}^{\text{L}}$*}         & \multicolumn{1}{r|}{11.09M}  & 68.9 & 8.71 & 46.1 & 71.3 & 2.47 \\
& \multicolumn{1}{l|}{$\text{Adpt}^{\text{H}}$*}         & \multicolumn{1}{r|}{11.09M}  & 67.3\textsubscript{$\pm$.6} & 8.5\textsubscript{$\pm$.07} & 46.0\textsubscript{$\pm$.2} & 70.7\textsubscript{$\pm$.2} & 2.44\textsubscript{$\pm$.01} \\
& \multicolumn{1}{l|}{LoRA}      & \multicolumn{1}{r|}{0.35M}   & 68.9\textsubscript{$\pm$.3} & 8.76\textsubscript{$\pm$.06} & 46.6\textsubscript{$\pm$.1} & 71.5\textsubscript{$\pm$.1} & \textbf{2.53}\textsubscript{$\pm$.03} \\
& \multicolumn{1}{l|}{\textbf{FourierFT}} & \multicolumn{1}{r|}{0.048M}   & \textbf{69.1}\textsubscript{$\pm$.1}  & \textbf{8.82} \textsubscript{$\pm$.05} & \textbf{47.0} \textsubscript{$\pm$.3} & \textbf{71.8} \textsubscript{$\pm$.1} & 2.51\textsubscript{$\pm$.02} \\ \midrule
\multirow{5}{*}{\rotatebox{90}{\begin{tabular}[c]{@{}c@{}} GPT-2\\ Large\end{tabular}}} & 
\multicolumn{1}{l|}{FT*}        & \multicolumn{1}{r|}{774.03M} & 68.5 & 8.78 & 46.0 & 69.9 & 2.45 \\
& \multicolumn{1}{l|}{$\text{Adpt}^{\text{L}}$*}         & \multicolumn{1}{r|}{0.88M}   & 69.1\textsubscript{$\pm$.1} & 8.68\textsubscript{$\pm$.03} & 46.3\textsubscript{$\pm$.0} & 71.4\textsubscript{$\pm$.2} & 2.49\textsubscript{$\pm$.0} \\
& \multicolumn{1}{l|}{$\text{Adpt}^{\text{L}}$*}         & \multicolumn{1}{r|}{23.00M}  & 68.9\textsubscript{$\pm$.3} & 8.70\textsubscript{$\pm$.04} & 46.1\textsubscript{$\pm$.1} & 71.3\textsubscript{$\pm$.2} & 2.45\textsubscript{$\pm$.02} \\
& \multicolumn{1}{l|}{LoRA}      & \multicolumn{1}{r|}{0.77M}   & 70.1\textsubscript{$\pm$.3} & 8.83\textsubscript{$\pm$.02} & 46.8\textsubscript{$\pm$.2} & \textbf{72.0}\textsubscript{$\pm$.3} & 2.47\textsubscript{$\pm$.02} \\
& \multicolumn{1}{l|}{\textbf{FourierFT}} & \multicolumn{1}{r|}{0.072M} & \textbf{70.2}\textsubscript{$\pm$.2} & \textbf{8.90}\textsubscript{$\pm$.02} & \textbf{47.0}\textsubscript{$\pm$.2} & 71.8\textsubscript{$\pm$.1} &  \textbf{2.50} \textsubscript{$\pm$.02} \\ \bottomrule
\end{tabular}%
}
  
 \end{minipage}
 \quad
 \begin{minipage}[t]{0.395\textwidth}
  
  \caption{The average scores on MT-Bench and Vicuna assessed by GPT-4. $\dag$ indicates updating the layers other than \texttt{lm\_head}. Higher score is better.}
\label{tab:llama}
\addtolength{\tabcolsep}{-4.95pt}
\resizebox{\textwidth}{!}{%
\begin{tabular}{@{}l|l|r|cc@{}}
\toprule
Model & Method & \begin{tabular}[c]{@{}r@{}}\# Trainable\\ Parameters\end{tabular} & MT-Bench & Vicuna \\ \midrule
\multirow{3}{*}{LLaMA1-7B} & LoRA\dag      & 159.9M & 5.05\textsubscript{$\pm$.3} & \textbf{6.85}\textsubscript{$\pm$.4} \\ 
& LoRA      & 33.5M & 4.99\textsubscript{$\pm$.3} & 6.81\textsubscript{$\pm$.3} \\ 
                           & \textbf{FourierFT} & 0.064M & \textbf{5.09}\textsubscript{$\pm$.6} & \textbf{6.85}\textsubscript{$\pm$.8} \\ \midrule
\multirow{3}{*}{LLaMA1-13B} & LoRA\dag         & 250.3M & \textbf{5.28}\textsubscript{$\pm$.6} & 7.02\textsubscript{$\pm$.3} \\ 
& LoRA     & 52.4M & 5.21\textsubscript{$\pm$.4} & 6.97\textsubscript{$\pm$.4} \\
                           & \textbf{FourierFT} & 0.08M & 5.23\textsubscript{$\pm$.3}  & \textbf{7.14}\textsubscript{$\pm$.5} \\ \midrule
\multirow{3}{*}{LLaMA2-7B} & LoRA\dag      & 159.9M & 5.19\textsubscript{$\pm$.1} & 7.38\textsubscript{$\pm$.3} \\ 
& LoRA      & 33.5M & \textbf{5.20}\textsubscript{$\pm$.3} & 7.35\textsubscript{$\pm$.6} \\ 
                           & \textbf{FourierFT} & 0.064M & 5.18\textsubscript{$\pm$.3} &  \textbf{7.49}\textsubscript{$\pm$.4}  \\ \midrule
\multirow{3}{*}{LLaMA2-13B} & LoRA\dag         & 250.3M & 5.78\textsubscript{$\pm$.2} & 7.89\textsubscript{$\pm$.5} \\
& LoRA      & 52.4M & 5.80\textsubscript{$\pm$.2} & 7.89\textsubscript{$\pm$.6}  \\
                           & \textbf{FourierFT} & 0.08M & \textbf{5.82}\textsubscript{$\pm$.3}  & \textbf{7.92}\textsubscript{$\pm$.5} \\\bottomrule
\end{tabular}%
}

 \end{minipage}
 
\end{table*}
\subsection{Natural Language Understanding}
\paragraph{Models and Datasets.}We evaluate our method on the GLUE benchmark (General Language Understanding Evaluation~\cite{glue}), which consists of a wide range of natural language understanding~(NLU) tasks, including single-sentence classification tasks, similarity and paraphrase tasks and natural language inference tasks. We fine-tune the pre-trained RoBERTa Base and Large foundation models~\cite{roberta} for evaluation. 
\paragraph{Implementation Details.}For both models, FourierFT is allowed to have 1000 out of $768^2$ (RoBERTa Base) and $1024^2$ (RoBERTa Large) trainable spectral coefficients in each layer, i.e., $n=1000$. We randomly sample the spectral entries with no frequency bias, which is shared\footnote{We use the value $2024$ as the \texttt{seed} for all layers.} across all 24 (Base) and 48 (Large) layers. For all 6 datasets in GLUE, we tune the hyperparameters of the learning rates and the scaling values. We follow the experimental setup applied in \citet{lora}, which involves fine-tuning only the query and value weights in each transformer block and fully fine-tuning the classification head. We provide the hyperparameters in Table~\ref{tab:hyper-glue} in Appendix.

\paragraph{Results.}Results are summarized in Table~\ref{tab:glue}. Following~\citet{lora}, \citet{adalora} and \citet{dylora}, we specify the number of trainable parameters for the fine-tuned layers excluding the classification head. We report the median of 5 random seed results, where the best epoch is selected for each run. In general, FourierFT achieves better or on-par performance compared with baseline methods with significantly fewer trainable parameters. Notably, FourierFT outperforms all baselines including fully fine-tuning the RoBERTa Base on CoLA and the RoBERTa Large on RTE. As mentioned in Section~\ref{sec32}, the parameter count of LoRA is dependent on both the width and depth of models, resulting in a larger count growth (LoRA: $0.8\text{M}/0.3\text{M}\approx2.7$; ours: $0.048\text{M}/0.024\text{M}=2$) compared to FourierFT. Nevertheless, FourierFT still performs comparably to LoRA, demonstrating the potential scalability of our method when facing even larger models.
\subsection{Natural Language Generation}
\paragraph{Models and Datasets.}
We evaluate the performance of FourierFT on the E2E natural language generation~(NLG) task~\cite{e2e}. We fine-tune the GPT-2~\cite{gpt2} Medium (354M) and Large (774M) models, which are both decoder-only and have 24 and 36 transformer blocks, respectively. The E2E benchmark contains roughly 42,000 training, 4,600 validation and 4,600 test samples from the restaurant domain.
\paragraph{Implementation Details.}We report prior results for baselines other than LoRA. For both LoRA and our method, we fine-tune the GPT-2 Medium and Large models with a linear learning rate scheduler for 5 epochs, where we tune the batch size and learning rate. We report the average results over 3 runs, where the last epoch is selected for each run. We provide the hyperparameters in Table~\ref{tab:hyper-e2e} in Appendix.
\paragraph{Results.}We show the results in Table~\ref{tab:e2e}. We note that FourierFT can achieve the best performance on most metrics. More importantly, FourierFT only requires $13.7\%$ and $9.4\%$ of the parameter counts of LoRA, for the GPT-2 Medium and Large models respectively. 

\begin{table*}[t!]
\centering
\caption{Fine-tuning results with ViT Base and Large models on different image classification datasets. We report the accuracy (\%) after 10 epochs. Avg. represents the average accuracy of each method on all datasets. The best performance is shown in \textbf{bold}.}
\label{tab:cv}
\addtolength{\tabcolsep}{-4.3pt}
\resizebox{0.99\textwidth}{!}{%
\begin{tabular}{@{}c|l|r|ccccccccc@{}}
\toprule
Model & Method & \begin{tabular}[c]{@{}r@{}}\# Trainable\\ Parameters\end{tabular} & \textbf{OxfordPets} & \textbf{StanfordCars} & \textbf{CIFAR10} & \textbf{DTD} & \textbf{EuroSAT} & \textbf{FGVC} & \textbf{RESISC45} & \textbf{CIFAR100} & \textbf{Avg.} \\ \midrule
\multirow{5}{*}{\rotatebox{90}{ViT-Base}} & LP & - & 90.28\textsubscript{$\pm$0.43} & 25.76\textsubscript{$\pm$0.28} & 96.41\textsubscript{$\pm$0.02} & 69.77\textsubscript{$\pm$0.67} & 88.72\textsubscript{$\pm$0.13} & 17.44\textsubscript{$\pm$0.43} & 74.22\textsubscript{$\pm$0.10} & 84.28\textsubscript{$\pm$0.11} & 68.36\\
 & FF & 85.8M & 93.14\textsubscript{$\pm$0.40} & \textbf{79.78}\textsubscript{$\pm$1.15} & \textbf{98.92}\textsubscript{$\pm$0.05} & \textbf{77.68}\textsubscript{$\pm$1.21} & \textbf{99.05}\textsubscript{$\pm$0.09} & \textbf{54.84}\textsubscript{$\pm$1.23} & \textbf{96.13}\textsubscript{$\pm$0.13} & \textbf{92.38}\textsubscript{$\pm$0.13} & \textbf{86.49}\\
 & LoRA & 581K & 93.19\textsubscript{$\pm$0.36} & 45.38\textsubscript{$\pm$0.41} & 98.78\textsubscript{$\pm$0.05} & 74.95\textsubscript{$\pm$0.40} & 98.44\textsubscript{$\pm$0.15} & 25.16\textsubscript{$\pm$0.16} & 92.70\textsubscript{$\pm$0.18} & 92.02\textsubscript{$\pm$0.12} & 77.58\\
 & \textbf{FourierFT} & 72K & \textbf{93.21}\textsubscript{$\pm$0.26} & 46.11\textsubscript{$\pm$0.24} & 98.58\textsubscript{$\pm$0.07} & 75.09\textsubscript{$\pm$0.37} & 98.29\textsubscript{$\pm$0.04} & 27.51\textsubscript{$\pm$0.64} & 91.97\textsubscript{$\pm$0.31} & 91.20\textsubscript{$\pm$0.14} & 77.75\\
 & \textbf{FourierFT} & 239K & 93.05\textsubscript{$\pm$0.34} & 56.36\textsubscript{$\pm$0.66} & 98.69\textsubscript{$\pm$0.08} & 77.30\textsubscript{$\pm$0.61} & 98.78\textsubscript{$\pm$0.11} & 32.44\textsubscript{$\pm$0.99} & 94.26\textsubscript{$\pm$0.20} & 91.45\textsubscript{$\pm$0.18} & 80.29\\ \midrule
\multirow{5}{*}{\rotatebox{90}{ViT-Large}} & LP & - &  91.11\textsubscript{$\pm$0.30} & 37.91\textsubscript{$\pm$0.27} & 97.78\textsubscript{$\pm$0.04} & 73.33\textsubscript{$\pm$0.26} & 92.64\textsubscript{$\pm$0.08} & 24.62\textsubscript{$\pm$0.24} & 82.02\textsubscript{$\pm$0.11} & 
84.28\textsubscript{$\pm$0.11} & 72.96\\
 & FF & 303.3M & 94.43\textsubscript{$\pm$0.56} & \textbf{88.90}\textsubscript{$\pm$0.26} & \textbf{99.15}\textsubscript{$\pm$0.05} & 81.79\textsubscript{$\pm$1.01} & \textbf{99.04}\textsubscript{$\pm$0.08} & \textbf{68.25}\textsubscript{$\pm$1.63} & \textbf{96.43}\textsubscript{$\pm$0.07} & 93.58\textsubscript{$\pm$0.19} & \textbf{90.20} \\
 & LoRA & 1.57M & 94.82\textsubscript{$\pm$0.09} & 73.25\textsubscript{$\pm$0.36} & 99.13\textsubscript{$\pm$0.03} & 81.79\textsubscript{$\pm$0.45} & 98.63\textsubscript{$\pm$0.07} & 42.32\textsubscript{$\pm$0.98} & 94.71\textsubscript{$\pm$0.25} & \textbf{94.87}\textsubscript{$\pm$0.10} & 84.94 \\
 & \textbf{FourierFT} & 144K & 94.46\textsubscript{$\pm$0.28} & 69.56\textsubscript{$\pm$0.30} & 99.10\textsubscript{$\pm$0.04} & 80.83\textsubscript{$\pm$0.43} & 98.65\textsubscript{$\pm$0.09} & 39.92\textsubscript{$\pm$0.68} & 93.86\textsubscript{$\pm$0.14} & 93.31\textsubscript{$\pm$0.09} & 83.71\\
 & \textbf{FourierFT} & 480K & \textbf{94.84}\textsubscript{$\pm$0.05} & 79.14\textsubscript{$\pm$0.67} & 99.08\textsubscript{$\pm$0.01} & \textbf{81.88}\textsubscript{$\pm$0.50} & 98.66\textsubscript{$\pm$0.03} & 51.28\textsubscript{$\pm$0.68} & 95.20\textsubscript{$\pm$0.07} & 93.37\textsubscript{$\pm$0.11} & 86.68\\ 
\bottomrule
\end{tabular}%
}
\end{table*}
\subsection{Instruction Tuning}
\paragraph{Models and Datasets.}
Instruction tuning, as described in \cite{it1, it2, it3}, refers to the process of fine-tuning a language model on a collection of paired prompts and responses. We apply LoRA and FourierFT to fine-tune the LLaMA~\cite{llama} and LLaMA2~\cite{llama2} families. Specifically, we consider the LLaMA-7B, LLaMA-13B, LLaMA2-7B and LLaMA2-13B as base models, which are fine-tuned on the Alpaca dataset~\cite{alpaca}. Alpaca contains 51K instruction-following demonstrations generated from \texttt{text-davinci-003}~(GPT-3.5)~\cite{gpt35}. For evaluation, we use the fine-tuned models to generate responses for the pre-defined questions, which are from the MT-Bench~\cite{mtbench} and Vicuna Eval~\cite{vicuna}. GPT-4 takes these answers as input and evaluates them with scores within 10.

\paragraph{Implementation Details.}For LoRA, we use $r=64$ and apply two configurations: (1) updating all linear layers except the language modelling head~(\texttt{lm\_head}); (2) updating only the $W_Q$ and $W_V$ matrices. For FourierFT, we only adopt the latter configuration with $n=1000$. To ensure the feasibility of training on a single GPU, we deploy the quantization method in~\citet{qlora} for fine-tuning. We train with both methods for only one epoch, and report the average scores of all answers. We provide the hyperparameter setup in Table~\ref{tab:hyper-llama} in the Appendix.

\paragraph{Results.}The results are shown in Table~\ref{tab:llama}. We find that the expressive power of the 13B model is much stronger than that of the 7B model, regardless of which fine-tuning method is used. Moreover, FourierFT closely matches or slightly exceeds LoRA's performance with less than $0.2\%$ of its parameters. We provide practical examples containing questions, answers and reviews in the Appendix~\ref{app:example}.

\subsection{Image Classification}
\paragraph{Models and Datasets.} We evaluate our method on the image classification task. We adopt the Base and Large versions of the popular CV foundation model, Vision Transformer (ViT)~\cite{vit}. The ViTs are pre-trained on the ImageNet-21K dataset \cite{ridnik2021imagenet}. The datasets for fine-tuning include OxfordPets (37\footnote{Numbers in parentheses indicate class counts for each dataset.}), CIFAR10 (10), DTD (47), EuroSAT (10) and RESISC45 (45) with small label spaces, as well as StanfordCars (196), FGVC (100) and CIFAR100 (100) with large label spaces. Detailed information is provided in Table \ref{tab:data-cv} in the Appendix.
\paragraph{Implementation Details.} We include three baselines for evaluation: Full Fine-tuning (FF), Linear Probing (LP, fine-tuning the classification head only), and LoRA. For both LoRA and our method, only the query and value matrices of ViT are updated. We use $r=16$ for LoRA and $n=\{3000, 10000\}$ for FourierFT. We tune the learning rates and weight decay for all methods, and set the maximum training epoch to 10. We provide the hyperparameters in Table~\ref{tab:hyper-cv} in Appendix.
\paragraph{Results.}Table~\ref{tab:cv} summarizes the results for 8 image classification datasets with the ViT Base and Large models. Both LoRA and FourierFT methods significantly outperform the Linear Probing, demonstrating their effectiveness in the CV domain. Our method obtains matched performance using 12.4\% and 9.2\% of LoRA's parameter count, with ViT Base and Large models, respectively. Notably, when we increase the parameter count of FourierFT to 41.1\% (ViT Base) and 30.6\% (ViT Large) of LoRA's, it can outperform LoRA by 3.5\% and 2.0\% respectively. Moreover, our method can even (slightly) outperform the Full Fine-tuning method on OxfordPets and DTD with the ViT Large model.

\begin{figure*}[t]
\centering
\includegraphics[width=0.89\textwidth]{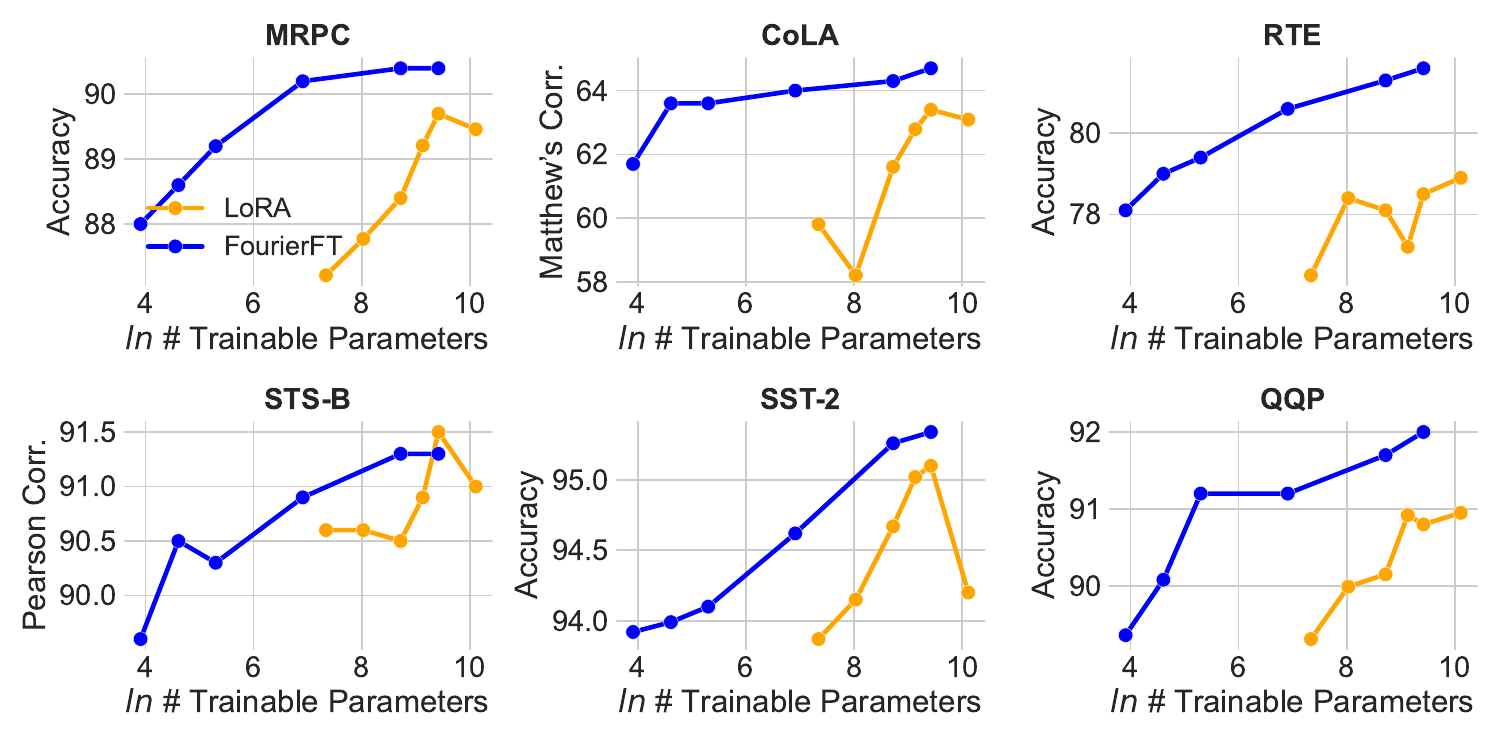} 
\caption{Performance on the GLUE benchmark with RoBERTa Base \textbf{vs.} number of trainable parameters (each layer) of LoRA and ours. For all 6 datasets, we apply the setting of $r=\{1,2,4,6,8,15\}$ for LoRA and $n=\{50,100,200,1000,6144,12288\}$.}
\label{fig:scale}
\end{figure*}

\subsection{Study}

\begin{figure}[t]
\centering
\includegraphics[width=0.47\textwidth]{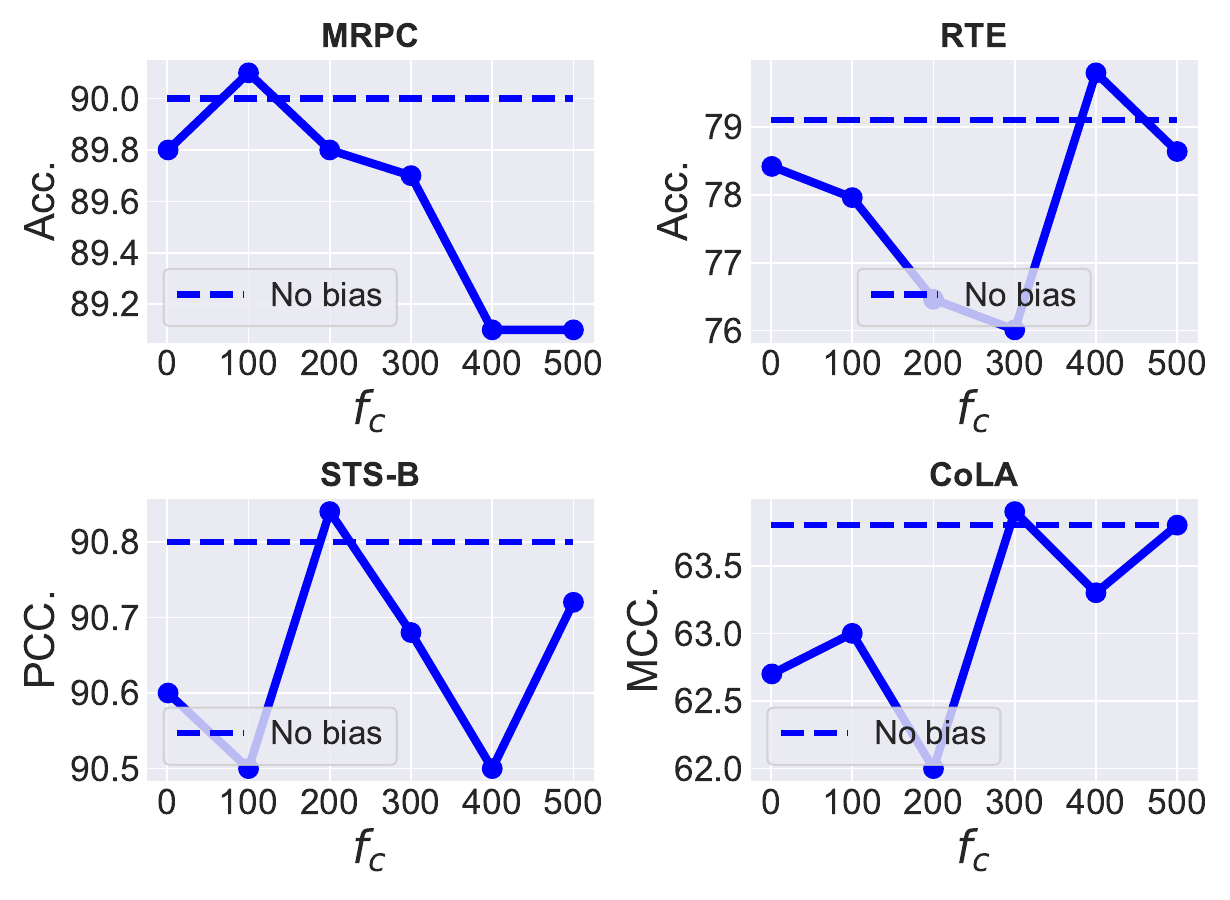} 
\caption{Results on 4 datasets in GLUE with different $f_c$ values.}
\label{fig:bias-main}
\vspace{-1.em}
\end{figure}

\paragraph{Effect of Frequency Bias.}We examine how the performance is affected by the frequency bias, i.e., the central frequency $f_c$ in Eq.~\ref{eq5}. We directly apply the optimal hyperparameters searched in Table~\ref{tab:glue} and fine-tune the RoBERTa Base on the MRPC, STS-B, CoLA and RTE datasets. From Figure~\ref{fig:bias-main}, we note that the fine-tuning performance of FourierFT without any frequency bias can surpass most cases that are restricted by the central frequency bias. This indicates the universality of our method. Surprisingly, we find that it is always possible to obtain results better than ``No bias" by traversing the $f_c$ values. Since this traversal is not efficient, we do not conduct further exploration in this paper. However, we believe that making $f_c$ trainable will be a promising new direction for improving FourierFT.

\paragraph{Parameter Scalability.}
We explore the relationship between the number of trainable parameters and the performance of LoRA and our method. We use the set of ranks $r=\{1,2,4,6,8,15\}$ for LoRA and $n=\{50,100,200,1000,6144,12288\}$ for FourierFT on 6 tasks of the GLUE benchmark. For both LoRA and ours, the learning rate, and scaling hyperparameters are tuned. For fairness, we ensure that the number of trials for hyperparameter search is 30 for both methods. As shown in Figure~\ref{fig:scale}, our method outperforms LoRA on all 6 datasets. In detail, our method is significantly better than LoRA with the same parameter count, i.e., $\{r=4, n=6144\}$ \& $\{r=8, n=12288\}$. Moreover, we observe that a larger number of parameters does not always bring performance gains for LoRA. On the contrary, the increase of $n$ can consistently improve the accuracy of FourierFT. On most tasks, FourierFT with $n=50$ can achieve comparable or even better (MRPC, CoLA, RTE) performance than LoRA with $r=1$. In this case, the parameter count in LoRA is about 31 $\times$ that of ours.

\paragraph{Basis Expressiveness.}The inverse discrete Fourier transform (IDFT) in Eq.~\ref{eq3} is equivalent to the matrix multiplication~\cite{tpu}: $S = \mathcal{B}_fF\mathcal{B}_f^{\top}$, where $\mathcal{B_f}$ is the transformation matrix of IDFT that contains the Fourier basis. To evaluate its expressivity, we replace the Fourier basis with random and orthogonal basis, respectively. Specifically, for $F\in\mathbb{R}^{d_1\times d_2}$, we initialize random basis $\mathcal{B}_{r}^1\in\mathbb{R}^{d_1\times d_1}$ and $\mathcal{B}_{r}^2\in\mathbb{R}^{d_2\times d_2}$ with the normal Gaussian distribution. Then Eq.~\ref{eq3} becomes $S = \mathcal{B}_{r}^1F\mathcal{B}_{r}^2$. 
A similar way is used for the orthogonal basis.
We compare FourierFT with the random basis (R-B) and orthogonal basis (O-B) on the GLUE benchmark. Table~\ref{tab:basis} shows the results. We note that the Fourier basis used in our method outperforms the random and orthogonal basis. In addition, the expressive power of the orthogonal basis is much stronger than that of the random basis. The stronger expressive power of the Fourier basis compared to the general orthogonal basis may be attributed to its effective capture of the spectral information of $\Delta W$.
\begin{table}[h!]
\centering
\caption{Results with three types of basis.}
\label{tab:basis}
\addtolength{\tabcolsep}{-3.5pt}
\resizebox{0.48\textwidth}{!}{%
\begin{tabular}{@{}l|ccc|ccc@{}}
\toprule
\multirow{2}{*}{Model} & \multicolumn{3}{c|}{RTE} & \multicolumn{3}{c}{CoLA} \\ \cmidrule(l){2-7} 
 & \textbf{Ours} & R-B & O-B & \textbf{Ours} & R-B & O-B \\ \midrule
Base & \textbf{79.1} & 72.7($\downarrow$8.1\%) & 75.6($\downarrow$4.4\%) & \textbf{63.8} & 58.7($\downarrow$8.0\%) & 60.0($\downarrow$6.0\%) \\
Large & \textbf{87.4} & 81.8($\downarrow$6.4\%) & 83.6($\downarrow$4.3\%) & \textbf{67.1} & 64.8($\downarrow$3.4\%) & 66.1($\downarrow$1.5\%) \\ \bottomrule
\end{tabular}%
}
\end{table}

\section{Conclusion}
In this paper, we aim to achieve an extremely low storage memory for a single fine-tuning of large foundation models. This will enable the customization of multiple fine-tunings for different domains, tasks, or user preferences. To achieve this, we propose a simple yet powerful fine-tuning method that treats weight changes as spatial-domain matrices and only learns the sparse coefficients in the spectral domain. Compared to the LoRA-style baselines, our approach reduces the number of trainable parameters by about $8\sim500\times$ on a wide range of tasks in the NLP and CV domains.

\section{Impact Statements}
This paper presents a work whose goal is to advance the field of Machine Learning. There are many potential societal consequences of our work, none which we feel must be specifically highlighted here.
\section*{Acknowledgements}
This work was supported by NSFC Grant No.62206067, HKUST--HKUST(GZ) 20 for 20 Cross-campus Collaborative Research Scheme C019 and Guangzhou-HKUST(GZ) Joint Funding Scheme 2023A03J0673.

\bibliography{main}
\bibliographystyle{icml2024}

\appendix
\onecolumn
\icmltitle{Supplementary of ``Parameter-Efficient Fine-Tuning \\ with Discrete Fourier Transform''}
\section{Details of Datasets}
\subsection{GLUE Benchmark}
The GLUE~\cite{glue} (General Language Understanding Evaluation) benchmark is widely used in the NLP domain. GLUE consists of a set of 8 NLP datasets: MNLI(inference), SST-2 (sentiment analysis), MRPC (paraphrase detection), CoLA (linguistic acceptability), QNLI (inference), QQP (question-answering), RTE (inference), and STS-B (textual similarity). We summarise their statistics in the following table.
\begin{table*}[h!]
\centering
\caption{Task descriptions and dataset statistics of the GLUE benchmark. STS-B belongs to the regression task. All other tasks are single sentence or sentence pair classification tasks.}
\label{tab:data-glue}
\resizebox{0.9\textwidth}{!}{%
\begin{tabular}{@{}llrrrrll@{}}
\toprule
\multicolumn{1}{l|}{\textbf{Corpus}} & Task & \# Train & \# Val & \# Test & \# Labels & Metrics & Domain \\ \midrule
\multicolumn{8}{c}{Single-Sentence Tasks} \\ \midrule
\multicolumn{1}{l|}{CoLA} & Acceptability & 8.55k & 1.04k & 1.06k & 2 & Matthews Corr. & misc. \\
\multicolumn{1}{l|}{SST-2} & Sentiment & 67.3k & 872 & 1.82k & 2 & Accuracy & Movie reviews \\ \midrule
\multicolumn{8}{c}{Similarity and Paraphrase Tasks} \\ \midrule
\multicolumn{1}{l|}{MRPC} & Paraphrase & 3.67 & 408 & 1.73k & 2 & Accuracy/F1 & News \\
\multicolumn{1}{l|}{STS-B} & Sentence similarity & 5.75k & 1.5k & 1.38k & 1 & Pearson/Spearman Corr. & misc. \\
\multicolumn{1}{l|}{QQP} & Paraphrase & 364k & 40.4k & 391k & 2 & Accuracy/F1 & Social QA \\ \midrule
\multicolumn{8}{c}{Inference Tasks} \\ \midrule
\multicolumn{1}{l|}{MNLI} & NLI & 393k & 19.65k & 19.65k & 3 & Accuracy & misc. \\
\multicolumn{1}{l|}{QNLI} & QA/NLI & 105k & 5.46k & 5.46k & 2 & Accuracy & Wikipedia \\
\multicolumn{1}{l|}{RTE} & NLI & 2.49k & 277 & 3k & 2 & Accuracy & News \& Wikipedia \\ \bottomrule
\end{tabular}%
}
\end{table*}
\subsection{E2E Benchmark}
The E2E (End-to-End) NLG challenge, proposed by \cite{e2e}, is a dataset for evaluating natural language
(data-to-text) generation models. The E2E dataset contains about 42,000 training samples, 4,600 validation samples and 4,600 test samples from the restaurant domain. E2E involves evaluation on 5 metrics: BLEU, NIST, METEOR, ROUGE-L, and CIDEr. A more detailed explanation of them is as follows.
\begin{itemize}
    \item \textbf{BLEU} (Bilingual Evaluation Understudy) is a metric to evaluate the quality of machine-generated text by comparing it to one or more human-generated reference texts.
    \item \textbf{NIST} (National Institute of Standards and Technology) is a metric that evaluates the quality of machine-generated text, similar to BLEU. NIST uses a weighted average of n-gram precisions to calculate the final score, whereas BLEU uses a geometric average.
    \item \textbf{METEOR} (Metric for Evaluation of Translation with Explicit ORdering) aligns the words in the machine-generated text with their corresponding words in the reference text, and then calculates a score based on the harmonic mean of precision and recall. 
    \item \textbf{ROUGE} (Recall-Oriented Understudy for Gisting Evaluation) measures the longest common sub-sequence (LCS) between the machine-generated summary and the reference summary. It is particularly useful for evaluating summaries that contain paraphrases or rephrased sentences, as it considers the LCS rather than exact word overlap.
    \item \textbf{CIDEr} (Consensus-based Image Description) measures the similarity between the machine-generated captions and the human-generated captions by considering both the n-gram overlap and the consensus among human annotators. 
\end{itemize}

\subsection{Alpaca}
Alpaca is a newly proposed dataset that contains only the training set. Alpaca contains 51k instructions and demonstrations generated by the \texttt{text-davinci-003} model. It can be used to fine-tune language models for specific instructions and improve their ability to follow instructions accurately. A specific example is as follows.

\definecolor{codeblue}{rgb}{0.25,0.5,0.5}
    \lstset{
      basicstyle=\fontsize{9.2pt}{7.2pt}\ttfamily\bfseries,
      commentstyle=\fontsize{7.2pt}{7.2pt}\color{codeblue},
      keywordstyle=\fontsize{7.2pt}{7.2pt},
      frame=single,
    }
\begin{minipage}{16cm}
\begin{lstlisting}[language=python]
{
    "instructions": Transform the following sentence into the passive voice.
    "input": I bought a book.	
    "output": A book was bought by me.	
}
\end{lstlisting}
\end{minipage}

The \texttt{\textbf{instruction}} describes the target task which should be performed by the model. The \texttt{\textbf{input}} denotes optional context or input for the task. The \texttt{\textbf{output}} is the answer to the instruction generated by \texttt{text-davinci-003}.
\subsection{MT-bench and Vicuna}
\textbf{MT-bench}~\cite{mtbench} is a recently proposed benchmark containing a series of open-ended questions. These questions can evaluate the instruction-following ability of a language foundation model. MT-bench primarily distinguishes the abilities of many aspects of the models, including writing, roleplay, reasoning, math, coding, extraction, stem, and humanities. A specific example is as follows.

\definecolor{codeblue}{rgb}{0.25,0.5,0.5}
    \lstset{
      basicstyle=\fontsize{9.2pt}{7.2pt}\ttfamily\bfseries,
      commentstyle=\fontsize{7.2pt}{7.2pt}\color{codeblue},
      keywordstyle=\fontsize{7.2pt}{7.2pt},
      frame=single,
    }
\begin{minipage}{16cm}
\begin{lstlisting}[language=python]
{
    "Q1": The vertices of a triangle are at points (0, 0), (-1, 1), and (3, 3).
    What is the area of the triangle?
    "Q2(follow-up)": What is the area of the circle circumscribing the triangle?	
    "Solution": Q1. Area is 3. Q2. 5pi.	
}
\end{lstlisting}
\end{minipage}

\textbf{Vicuna Eval} refers to the benchmark for evaluating LLM alignment with human preferences, which is the predecessor of MT-bench. Vicuna Eval covers the topics of coding, writing, math, counterfactual, fermi, common sense, roleplay, knowledge, and generic. A specific example is as follows.

\definecolor{codeblue}{rgb}{0.25,0.5,0.5}
    \lstset{
      basicstyle=\fontsize{9.2pt}{7.2pt}\ttfamily\bfseries,
      commentstyle=\fontsize{7.2pt}{7.2pt}\color{codeblue},
      keywordstyle=\fontsize{7.2pt}{7.2pt},
      frame=single,
    }
\begin{minipage}{16cm}
\begin{lstlisting}[language=python]
{
    "question": Implement a binary search algorithm to find a specific element 
    in a sorted array.
    "category": coding.
}
\end{lstlisting}
\end{minipage}

\subsection{Image Classification Datasets}
The statistics of the selected 8 vision datasets are in Table~\ref{tab:data-cv}.

\begin{table}[h!]
\centering
\caption{Details about the vision datasets.}
\label{tab:data-cv}
\resizebox{0.7\textwidth}{!}{%
\begin{tabular}{@{}l|lrrrc@{}}
\toprule
Dataset & \#Train & \#Val & \#Test & \#Class & Rescaled resolution \\ \midrule
OxfordPets \cite{pets} & 3,312 & 368 & 3,669 & 37 & \multirow{8}{*}{$224\times224$} \\
StandfordCars \cite{cars} & 7,329 & 815 & 8,041 & 196 &  \\
CIFAR10 \cite{cifar} & 45,000 & 5,000 & 10,000 & 10 &  \\
DTD \cite{dtd}& 4,060 & 452 & 1,128 & 47 &  \\
EuroSAT \cite{euro} & 16,200 & 5,400 & 5,400 & 10 &  \\
FGVC \cite{fgvc} & 3,000 & 334 & 3,333 & 100 &  \\
RESISC45 \cite{resisc} & 18,900 & 6,300 & 6,300 & 45 &  \\
CIFAR100 \cite{cifar} & 45,000 & 5,000 & 10,000 & 100 &  \\ \bottomrule
\end{tabular}%
}
\end{table}
\clearpage
\section{Hyperparamaters}
\begin{table}[h!]
\centering
\caption{Hyperparameter setup of FourierFT for the GLUE benchmark.}
\label{tab:hyper-glue}
\resizebox{0.8\textwidth}{!}{%
\begin{tabular}{@{}clcccccc@{}}
\toprule
Model & Hyperparameter & \multicolumn{1}{|c}{STS-B} & \multicolumn{1}{c}{RTE} & \multicolumn{1}{c}{MRPC} & \multicolumn{1}{c}{CoLA} & \multicolumn{1}{c}{SST-2} & \multicolumn{1}{c}{QNLI} \\ \midrule
\multirow{5}{*}{Both} & \multicolumn{1}{l|}{Optimizer} & \multicolumn{6}{c}{AdamW} \\
 & \multicolumn{1}{l|}{LR Schedule} & \multicolumn{6}{c}{Linear} \\
 & \multicolumn{1}{l|}{Warmup Ratio} & \multicolumn{6}{c}{0.06} \\
 & \multicolumn{1}{l|}{Frequency Bias} & \multicolumn{6}{c}{False} \\
 & \multicolumn{1}{l|}{$n$} & \multicolumn{6}{c}{1000} \\
  & \multicolumn{1}{l|}{\texttt{seeds}} & \multicolumn{6}{c}{\{0, 11111, 22222, 33333, 44444\}} \\\midrule
\multirow{6}{*}{Base} & \multicolumn{1}{l|}{Epochs} & 60 & 90 & 30 & 100 & 40 & 40 \\
 & \multicolumn{1}{l|}{Learning Rate (FourierFT)} & 9E-2 & 9E-2 & 5E-2 & 1.2E-1 & 5E-2 & 1E-2 \\
 & \multicolumn{1}{l|}{Learning Rate (Head)} & 9E-3 & 1.1E-2 & 6E-3 & 8E-3 & 6E-3 & 1E-3 \\
 & \multicolumn{1}{l|}{Max Seq. Len} & 512 & 512 & 512 & 512 & 512 & 512 \\
 & \multicolumn{1}{l|}{Scaling value} & 84 & 110 & 141 & 49 & 140 & 29 \\
 & \multicolumn{1}{l|}{Batch Size} & 32 & 32 & 32 & 32 & 32 & 32 \\ \midrule
\multicolumn{1}{l}{\multirow{6}{*}{Large}} & \multicolumn{1}{l|}{Epochs} & 30 & 60 & 30 & 80 & 10 & 30 \\
\multicolumn{1}{l}{} & \multicolumn{1}{l|}{Learning Rate (FourierFT)} & 7E-2 & 8E-2 & 6E-2 & 4.3E-2 & 4.3E-2 & 6E-2 \\
\multicolumn{1}{l}{} & \multicolumn{1}{l|}{Learning Rate (Head)} & 1E-3 & 5E-3 & 1E-3 & 1.1E-2 & 1E-3 & 5E-3 \\
\multicolumn{1}{l}{} & \multicolumn{1}{l|}{Max Seq. Len} & 512 & 512 & 512 & 256 & 128 & 512 \\
\multicolumn{1}{l}{} & \multicolumn{1}{l|}{Scaling Value} & 121 & 90 & 120 & 120 & 99 & 69 \\
\multicolumn{1}{l}{} & \multicolumn{1}{l|}{Batch Size} & 32 & 32 & 32 & 128 & 32 & 32 \\ \bottomrule
\end{tabular}%
}
\end{table}
\begin{table}[h!]
\centering
\caption{Hyperparameter setup of FourierFT on the E2E benchmark.}
\label{tab:hyper-e2e}
\resizebox{0.4\textwidth}{!}{%
\begin{tabular}{@{}l|ll@{}}
\toprule
Hyperparameter & Medium & Large \\ \midrule
Optimizer & \multicolumn{2}{c}{AdamW} \\
Learning Rate (FourierFT) & 2E-2 & 5E-2 \\
Learning Rate (Head) & 2E-4 & 1E-4 \\
Batch Size & \multicolumn{2}{c}{128} \\
Weight Decay & 0.01 & 0.03 \\
$n$ & \multicolumn{2}{c}{1000} \\
Scaling value $\alpha$ & \multicolumn{2}{c}{300} \\
Epochs & \multicolumn{2}{c}{5} \\
Label Smooth &  \multicolumn{2}{c}{0.1}  \\
LR Schedule & \multicolumn{2}{c}{Linear}  \\ \bottomrule
\end{tabular}%
}
\end{table}
\begin{table}[h!]
\centering
\caption{Hyperparameter setup for instruction-tuning of LoRA and FourierFT.}
\label{tab:hyper-llama}
\resizebox{0.35\textwidth}{!}{%
\begin{tabular}{@{}l|lc@{}}
\toprule
Hyperparameter & LoRA & FourierFT \\ \midrule
Optimizer &  \multicolumn{2}{c}{AdamW}  \\
Warmup Ratio &  \multicolumn{2}{c}{0.06}  \\
Batch Size &  \multicolumn{2}{c}{4}  \\
Accumulation Steps &  \multicolumn{2}{c}{4}  \\
Epochs &  \multicolumn{2}{c}{1}  \\
$n$ & 1000 & -- \\
Scaling Value $\alpha$ & 300.0 & 16.0 \\
LR Schedule &  \multicolumn{2}{c}{Linear}  \\
Learning Rate & 3E-2 &  3E-3\\ \bottomrule
\end{tabular}%
}
\end{table}
\begin{table}[ht!]
\centering
\caption{Hyperparameter setup for image classification of FourierFT.}
\label{tab:hyper-cv}
\resizebox{0.9\textwidth}{!}{%
\begin{tabular}{@{}l|clllllll@{}}
\toprule
Hyperparameter & OxfordPets & \multicolumn{1}{c}{StanfordCars} & \multicolumn{1}{c}{CIFAR10} & \multicolumn{1}{c}{DTD} & \multicolumn{1}{c}{EuroSAT} & \multicolumn{1}{c}{FGVC} & \multicolumn{1}{c}{RESISC45} & \multicolumn{1}{c}{CIFAR100} \\ \midrule
Epochs & \multicolumn{8}{c}{10} \\
Optimizer & \multicolumn{8}{c}{AdamW} \\
LR Schedule & \multicolumn{8}{c}{Linear} \\
$n$ & \multicolumn{8}{c}{3000} \\
$\alpha$ & \multicolumn{8}{c}{300.0} \\
Learning Rate (FourierFT) & \multicolumn{1}{c}{3E-1} & \multicolumn{1}{c}{3E-1} & \multicolumn{1}{c}{3E-1} & \multicolumn{1}{c}{3E-1} & \multicolumn{1}{c}{2E-1} & \multicolumn{1}{c}{3E-1} & \multicolumn{1}{c}{3E-1} & \multicolumn{1}{c}{2E-1} \\
Learning Rate (Head) & \multicolumn{1}{c}{1E-3} & \multicolumn{1}{c}{1E-3} & \multicolumn{1}{c}{1E-3} & \multicolumn{1}{c}{1E-3} & \multicolumn{1}{c}{8E-4} & \multicolumn{1}{c}{1E-3} & \multicolumn{1}{c}{1E-3} & \multicolumn{1}{c}{7E-4} \\
Weight Decay & \multicolumn{1}{c}{8E-4} & \multicolumn{1}{c}{4E-5} & \multicolumn{1}{c}{9E-5} & \multicolumn{1}{c}{7E-5} & \multicolumn{1}{c}{3E-4} & \multicolumn{1}{c}{7E-5} & \multicolumn{1}{c}{3E-4} &  \multicolumn{1}{c}{1E-4} \\ \bottomrule
\end{tabular}%
}
\end{table}
\section{Additional Experimental Results}
\subsection{Training Curve}
\begin{figure*}[h]
\centering
\includegraphics[width=0.88\textwidth]{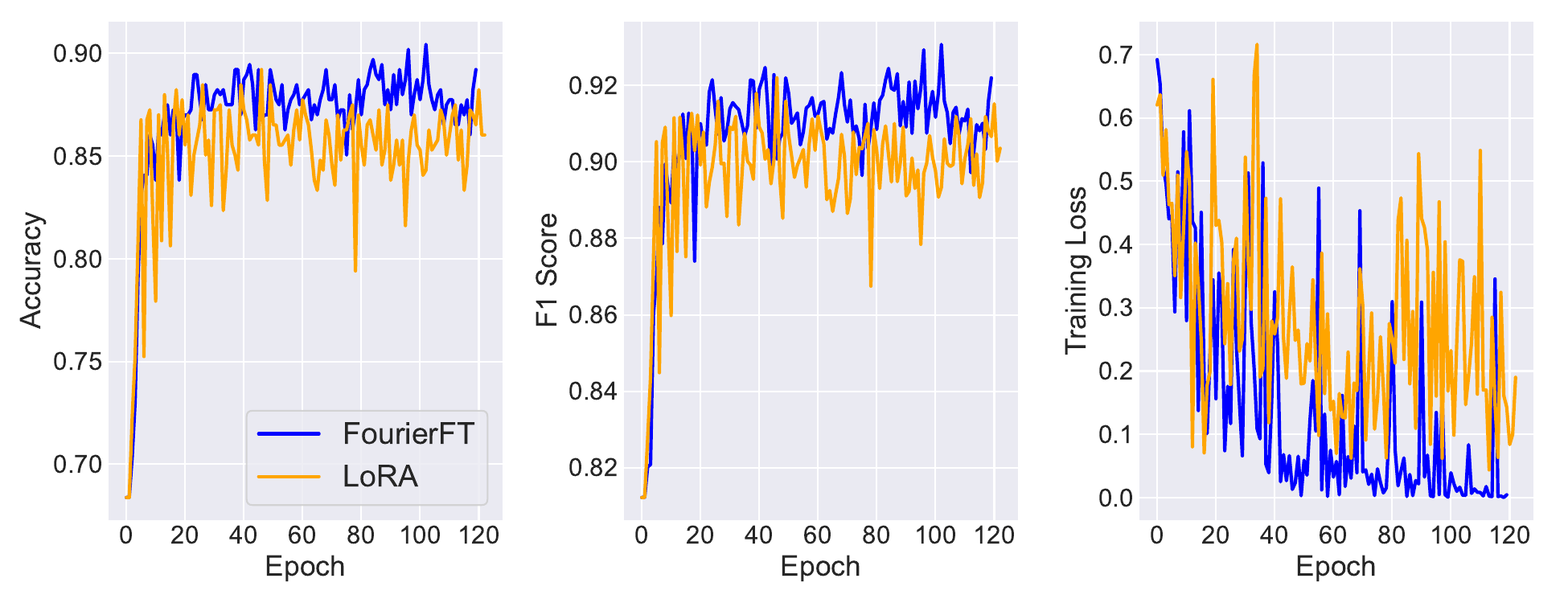} 
\caption{Training process of LoRA and ours. We show the current accuracy, F1 score and training loss of two methods.}
\label{fig:curve}
\end{figure*}

We show the training curves of our method and LoRA to demonstrate that the superior performance of FourierFT is not due to coincidence. In Figure \ref{fig:curve}, we set $r=1$ for LoRA and $n=1536$ for the MRPC task, so that the total number of trainable parameters is equivalent for both methods. It can be seen that FourierFT consistently outperforms LoRA in terms of accuracy, F1 score, and training loss throughout the entire training process.

\clearpage
\subsection{Expressive Ability}
\begin{figure*}[h]
\centering
\includegraphics[width=0.74\textwidth]{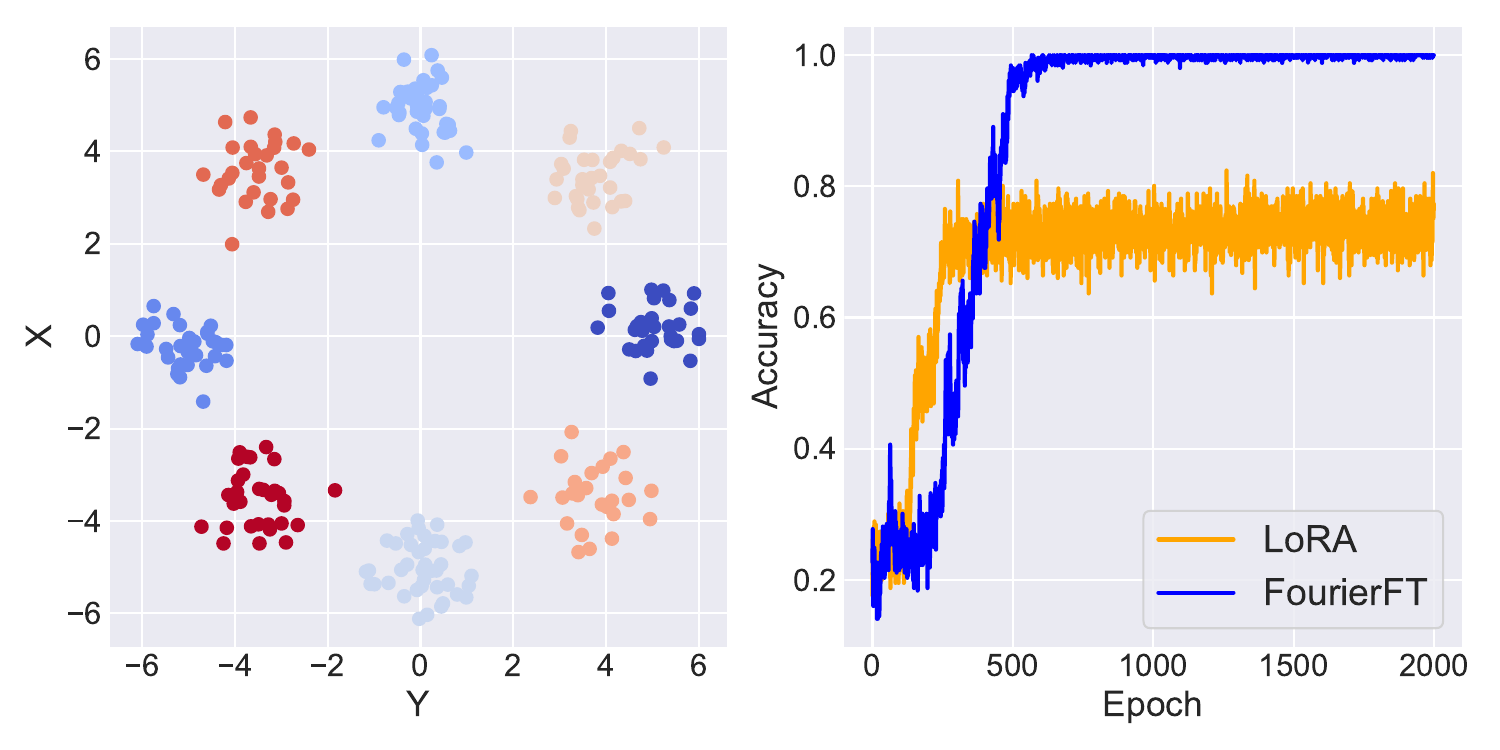} 
\caption{Left: The dataset distribution. Colors represent the classes. The model input is the 2D coordinate of each point. Right: Training process of LoRA and ours.}
\label{fig:renyi}
\end{figure*}

To intuitively evaluate the expressive power of our method, we design a simple classification task with a synthetic dataset to simulate a scenario where LoRA encounters performance bottlenecks. Specifically, we specify a 2D center point for each class of data in the 8 classes, and randomly add Gaussian noise based on that point to obtain the 2D coordinates of the input. The dataset visualization is shown on the left of Figure~\ref{fig:renyi}. We train a single $64*64$ sized hidden layer with LoRA ($r=1$) and FourierFT ($n=128$) to fit the synthesized data. In this case, both methods require the same number of trainable parameters. However, the results of the experiments are vastly different. It can be seen that LoRA never reaches 100\% accuracy within 2000 epochs, while FourierFT can quickly achieve it (in about 500 epochs). Under certain parameter constraints, LoRA has obvious performance bottlenecks, while FourierFT can easily overcome them.

\clearpage
\section{Examples of Instruction Tuning}\label{app:example}
We provide some practical cases generated by the LLaMA2-7B model that is fine-tuned with LoRA and ours. Each case includes the question, the answer and the review assigned by GPT-4.
\begin{figure*}[h!]
\centering
\includegraphics[width=0.95\textwidth]{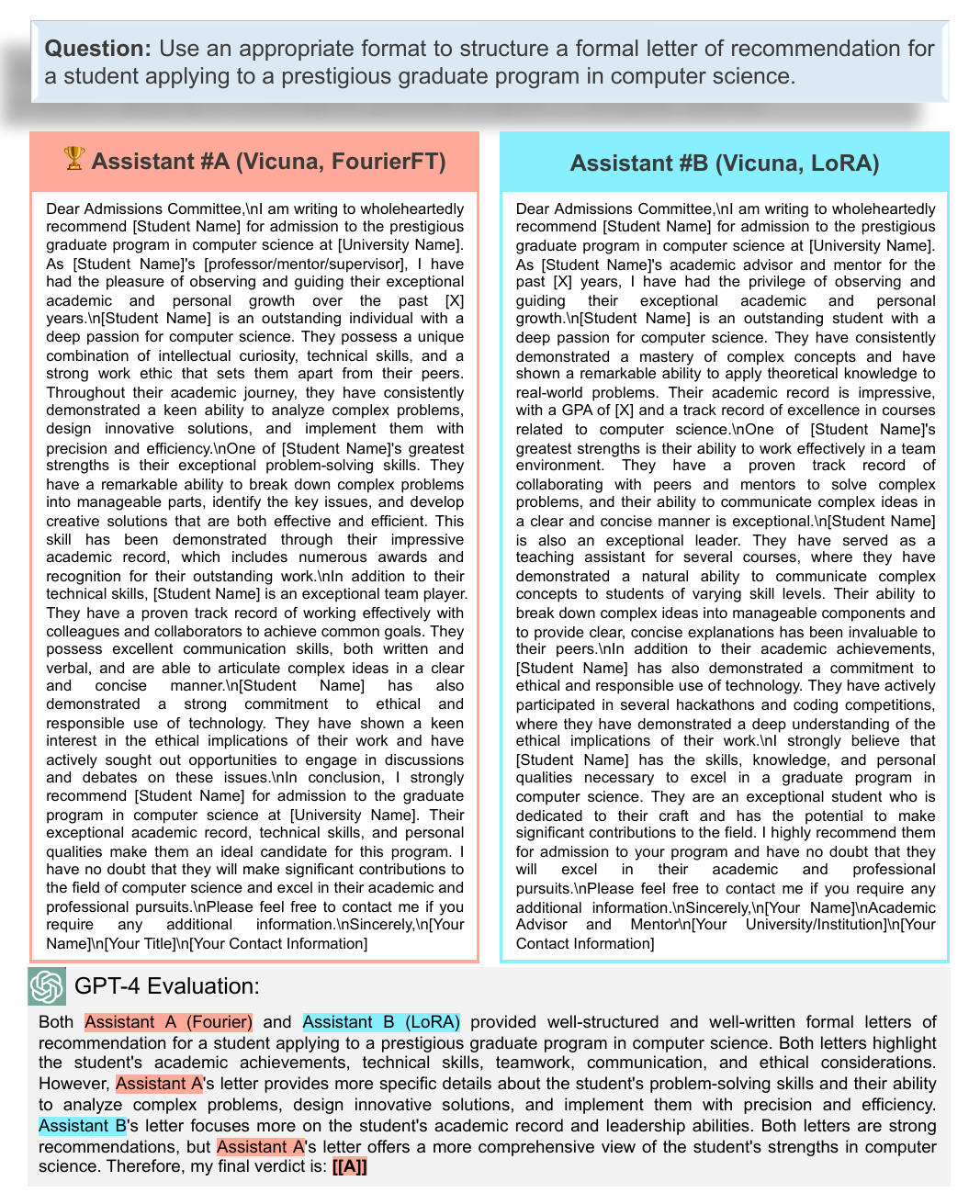} 
\caption{Example 1.}
\label{fig:example-1}
\end{figure*}
\begin{figure*}[h!]
\centering
\includegraphics[width=0.96\textwidth]{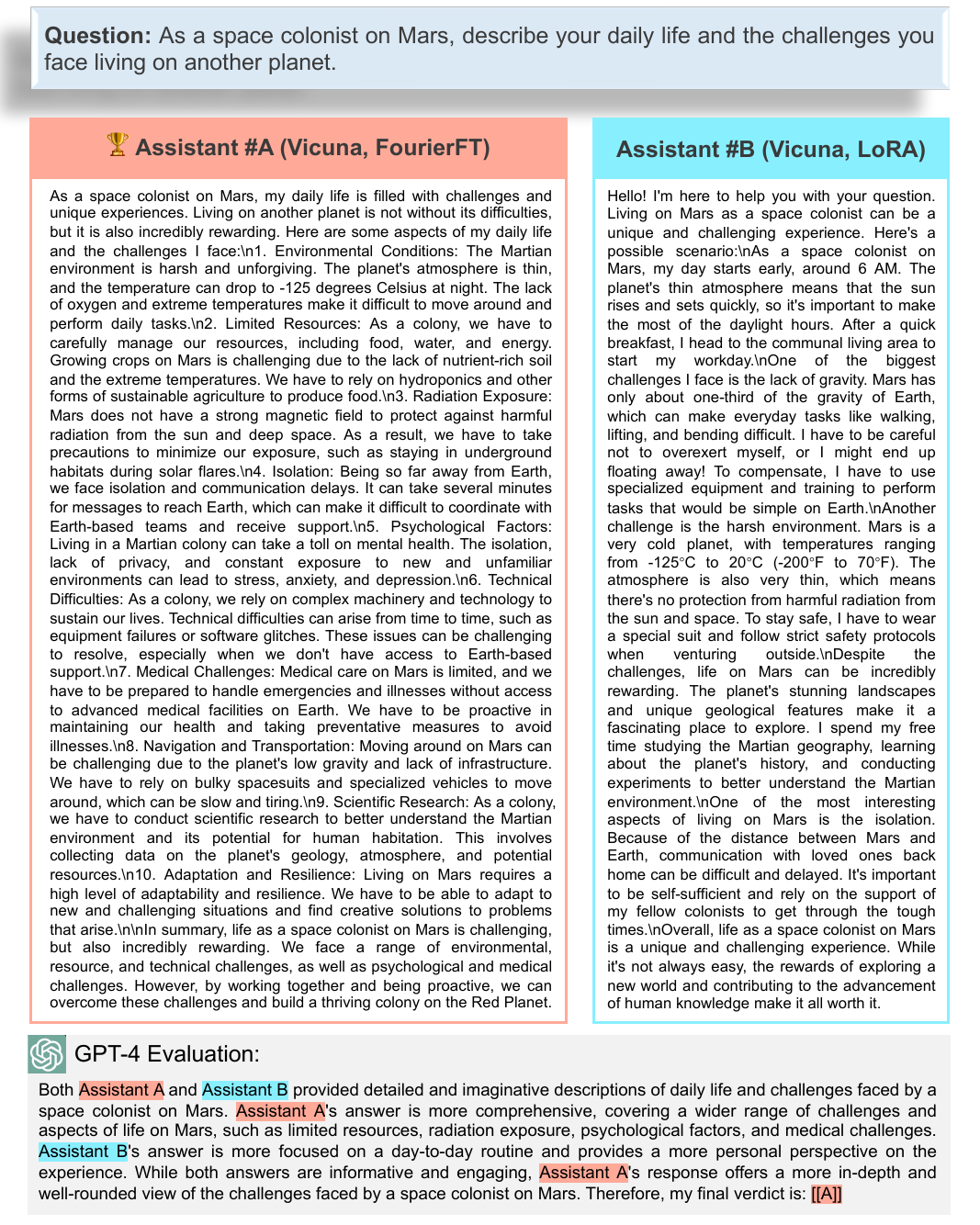} 
\caption{Example 2.}
\label{fig:example-2}
\end{figure*}

\section{Stable Diffusion fine-tuning: Dreambooth}
\paragraph{Models and Datasets.}
Following \citet{dreambooth}, we evaluate our method on the subject-driven text-to-image task, which refers to generating multiple images of a specified subject, guided by a textual prompt. We use Stable Diffusion v1.5 ~\cite{sd} (SD1.5) as the pre-trained text-to-image model, and compare our method with LoRA and DreamBooth. For fairness, we randomly pick generated images from LoRA and our method. For fine-tuning, we use the dataset proposed in Dreambooth \cite{dreambooth}, where five or six image samples are used for training for each subject. 

\paragraph{Implementation Details.} 
Both LoRA and our method use the same loss function as in DreamBooth. For DreamBooth, we apply the best hyperparameter setup in the original paper. For LoRA, we tune the rank $r$, the learning rate and the scaling value $\alpha$.

\begin{figure}[ht]
\centering
\includegraphics[width=0.99\textwidth]{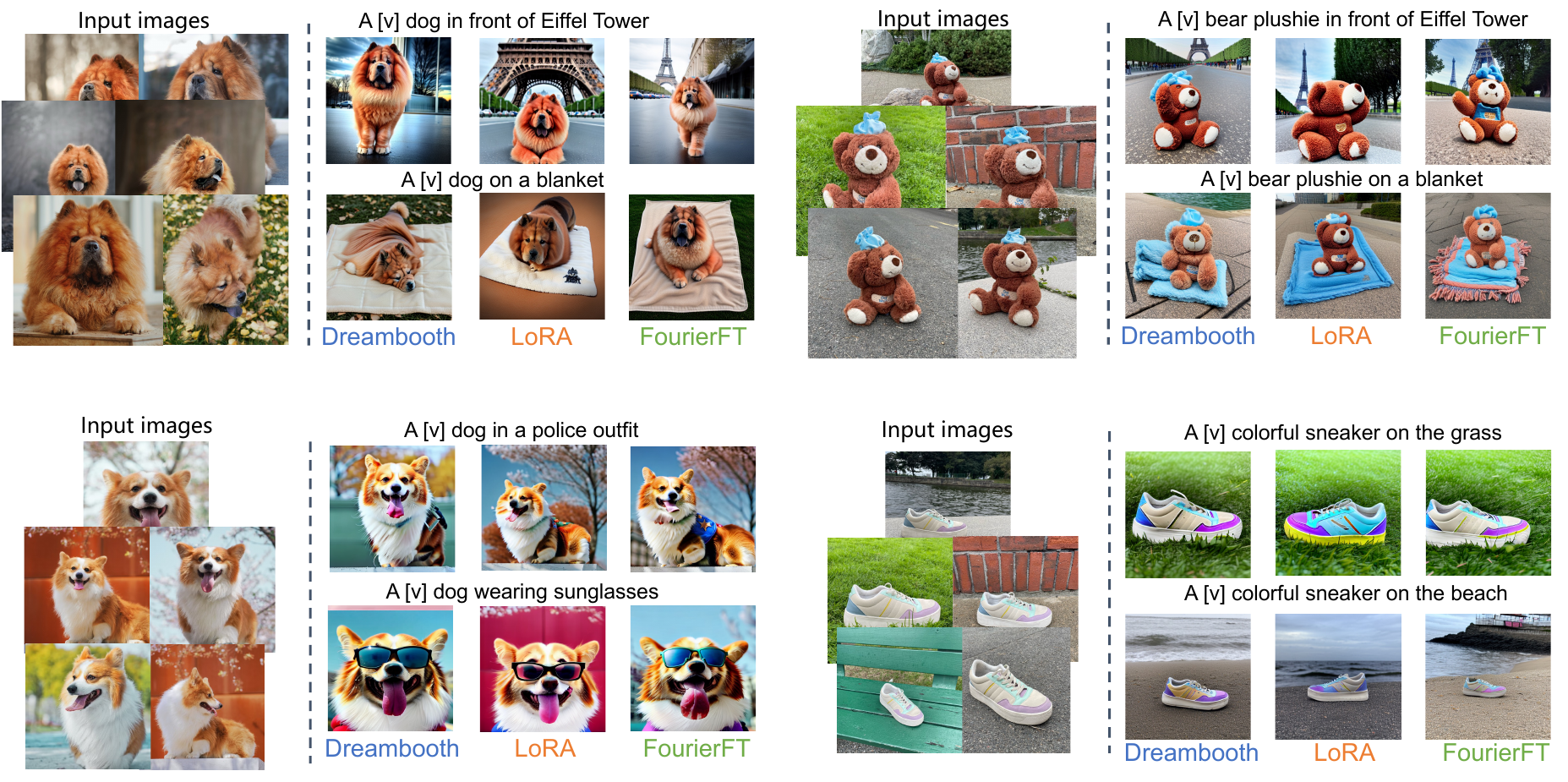} 
\vspace{-1.5em}
 \caption{Generated samples of DreamBooth, LoRA and FourierFT on the subject-driven generation task. All examples are randomly picked. The figure is best viewed digitally, in color and significantly zoomed in.}
\label{fig:sd1.5}
\end{figure}

\begin{table}[h!]
\centering
\caption{Results of fine-tuning methods on the FID metric, which measures the similarity between generated and target images.}
\label{tab:sd1.5}
\resizebox{0.4\textwidth}{!}{%
\begin{tabular}{@{}l|l|r|c@{}}
\toprule
Model & Method & \begin{tabular}[c]{@{}l@{}}\# Trainable\\ Parameters\end{tabular} & FID \\ \midrule
SD1.1 & DreamBooth & -- & 237.8 \\ \midrule
SD1.5 & W/o Fine-tuning & -- & 261.7 \\
SD1.5 & FF & 862 M & 221.6 \\
SD1.5 & LoRA & 12.4 M & 245.2 \\
SD1.5 & FourierFT & 0.19 M & 244.9 \\ \bottomrule
\end{tabular}%
}
\end{table}
\paragraph{Results.} 
Results are shown in Table \ref{tab:sd1.5}. Our method achieves comparable FID performance with only $6.1\%$ the parameters of LoRA's.

\end{document}